\documentclass[runningheads]{Sty_Ref/llncs}

\usepackage{graphicx}
\usepackage{amsmath}
\usepackage{amssymb}
\usepackage{booktabs}
\usepackage{comment}
\usepackage[accsupp]{axessibility}
\usepackage{wrapfig}

\usepackage{multirow}

\usepackage[colorlinks]{hyperref}
\usepackage{breakcites}
\newcommand\ZS[1]{{\textcolor{blue}{Zhixin: #1}}}

\usepackage[capitalize]{cleveref}
\crefname{section}{Sec.}{Secs.}
\Crefname{section}{Section}{Sections}
\Crefname{table}{Table}{Tables}
\crefname{table}{Tab.}{Tabs.}

\renewcommand\P{\mathbb{P}}
\newcommand\Real{\mathbb{R}}
\newcommand\Ours{3D-FM GAN}

\renewcommand{\L}{\mathcal{L}}

\newcommand\T{\mathcal{T}}
\newcommand\W{\mathcal{W}}

\newcommand\Nl{\mathbf{N_l}}
\newcommand\Nf{\mathbf{N_f}}
\newcommand\G{\mathbf{G}}
\newcommand\E{\mathbf{E}}
\newcommand\FR{\mathbf{FR}}
\newcommand\Rd{\mathbf{Rd}}
\newcommand\Gd{\mathbf{G_d}}

\newcommand\Revise{\color{red}}
\newcommand\Mark[1]{\textsuperscript#1}

\begin{document}

\pagestyle{headings}
\mainmatter
\def\ECCVSubNumber{1060}  % Insert your submission number here

\title{3D-FM GAN: Towards 3D-Controllable Face Manipulation} % Replace with your title

% INITIAL SUBMISSION 
\begin{comment}
\titlerunning{ECCV-22 submission ID \ECCVSubNumber} 
\authorrunning{ECCV-22 submission ID \ECCVSubNumber} 
\author{Anonymous ECCV submission}
\institute{Paper ID \ECCVSubNumber}
\end{comment}
%******************

% \author{Yuchen Liu\inst{1} \and Zhixin Shu\inst{2} \and Yijun Li\inst{2} \and Zhe Lin{2} \and Richard Zhang\inst{2} \and S.Y. Kung\inst{1}}

% \institute{Princeton University \and Adobe Research \\
% {\tt\small \Mark{1}\{yl16, kung\}@princeton.edu~~~\Mark{2}\{zshu, yijli, zlin, rizhang\}@adobe.com}
% }

\author{Yuchen Liu\Mark{1}\thanks{Work done during an internship at Adobe Research.}, Zhixin Shu\Mark{2}, Yijun Li\Mark{2}, Zhe Lin\Mark{2}, Richard Zhang\Mark{2}, S.Y. Kung\Mark{1}}

\institute{\Mark{1}Princeton University~~~\Mark{2}Adobe Research \\
{\tt\small \Mark{1}\{yl16, kung\}@princeton.edu~~~\Mark{2}\{zshu, yijli, zlin, rizhang\}@adobe.com}
}

\authorrunning{Y. Liu et al.}

\maketitle

%%%%%%%%% BODY TEXT
%%%%%%%%% ABSTRACT

\begin{abstract}
3D-controllable portrait synthesis has significantly advanced, thanks to breakthroughs in generative adversarial networks (GANs). 
However, it is still challenging to manipulate existing face images with precise 3D control.
While concatenating GAN inversion and a 3D-aware, noise-to-image GAN is a straight-forward solution, 
it is inefficient and may lead to noticeable drop in editing quality.
To fill this gap, we propose \Ours, a novel conditional GAN framework designed specifically for \textbf{3D}-controllable \textbf{F}ace \textbf{M}anipulation, 
and does not require any tuning after the end-to-end learning phase.
By carefully encoding both the input face image and a physically-based rendering of 3D edits into a StyleGAN's latent spaces, 
our image generator provides high-quality, identity-preserved, 3D-controllable face manipulation.
To effectively learn such novel framework,
we develop two essential training strategies
and a novel multiplicative co-modulation architecture that improves significantly upon naive schemes.
With extensive evaluations, we show that our method outperforms the prior arts on various tasks, with better editability, 
stronger identity preservation, and higher photo-realism. 
In addition, we demonstrate a better generalizability of our design on large pose editing and out-of-domain images.
More can be found in \href{https://lychenyoko.github.io/3D-FM-GAN-Webpage/}{webpage} and  \href{https://youtu.be/3tR7qIXyzLE}{video}.

\begin{comment}
\ZS{Thanks to breakthroughs in generative adversarial networks (GANs), high-quality portrait image generation has significantly advanced. 
However, it is still challenging to manipulate face images with precise 3D control and faithful identity preservation: existing 3D-controllable face generation work requires a two-step solution for image editing, and the results are often unsatisfactory. To address this, in this paper, we propose \Ours, a novel conditional StyleGAN architecture designed for \textbf{3D}-controllable \textbf{F}ace \textbf{M}anipulation. 
By carefully encoding both the input face image and a physically-based rendering of 3D edits into a StyleGAN's latent spaces, 
our image generator learns high-quality, identity-preserved, 3D-controllable face manipulation.
With extensive quantitative and qualitative evaluations, we show that our method outperforms the prior arts on various tasks, with better editability, stronger identity preservation, and higher photo-realism. 
In addition, we demonstrate better generalizability of our design on large pose editing and out-of-domain images.}
\end{comment}

\end{abstract}

\begin{figure}[t]
    \centering
  \includegraphics[width=\textwidth]{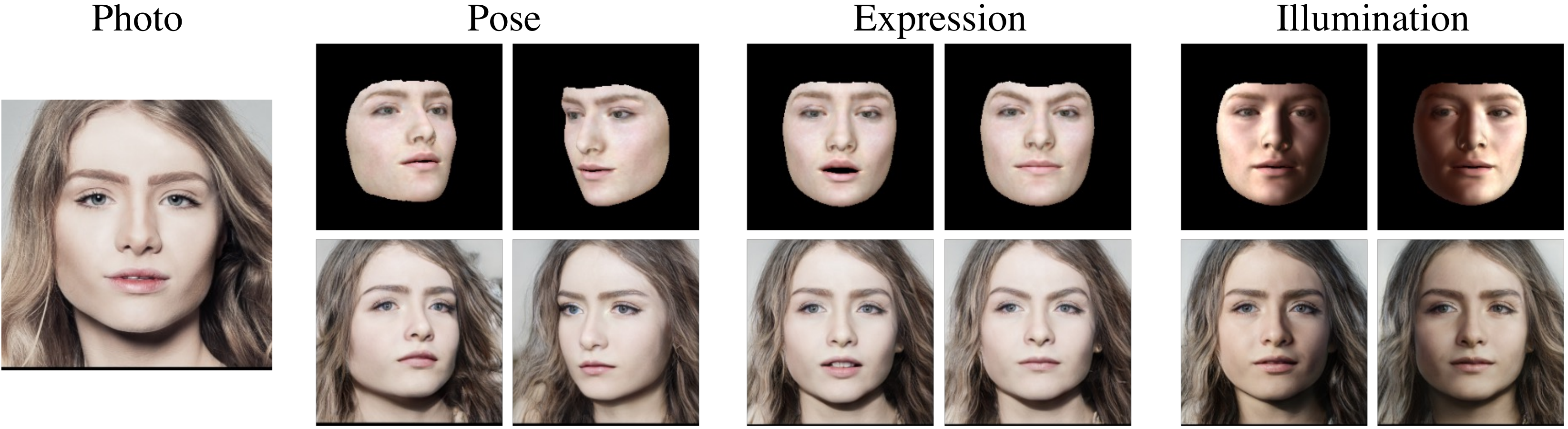}
    \vspace{-0.8cm}
    \caption{With explicit 3D controls of pose, expression, and illumination, presented as identity-preserved rendered faces (top row),
\Ours\ provides controllable and disentangled face manipulations on real world
images (bottom row) with strong identity preservation and high photo-realism.
    }
    \label{fig:teaser}
    \vspace{-0.6cm}
\end{figure}

\section{Introduction}
\label{sec:intro}

% P1: Face synthesis is an important problem and has many application. 

%P2: The drawback of conventional approach of graphics rendering 

Face manipulation with precise control has long attracted attention from computer vision and computer graphics community 
for its application in face recognition, photo editing, visual effects, and AR/VR applications, etc. 
In the past, researchers developed 3D morphable face models (3DMMs)~\cite{blanz1999morphable,paysan20093d,li2017learning},
which provide an explainable and disentangled parameter space to control face attributes of identity, pose, expression, and illumination.
However, it is still challenging to render photo-realistic face manipulations with 3DMMs.

%P3: Recent learning approach shows promising progress, however, in-controllability.

In recent years, generative adversarial networks (GANs)~\cite{goodfellow2014generative} have demonstrated promising results in photo-realistic face synthesis~\cite{karras2019style,karras2020analyzing} by mapping random noise to image domain.
While latent space exploration has been attempted~\cite{harkonen2020ganspace,abdal2019image2stylegan,shen2021closed}, 
it requires a lot of human labor to discover meaningful directions, 
and the editings could still be entangled. 
As such, the variants of conditional GANs are widely studied for identity-preserved face manipulations~\cite{tran2017disentangled,zhou2019deep,choi2018stargan,bao2018towards}.
Nonetheless, they either only allow control on a single facial attribute or require reference images/human annotations for face editing. 

%P4: Motivate to combine both techniques together, bringing the merits from both sides
%P5: Prior works and their drawbacks and discuss our advantages
More recently, several works introduced 3D priors into GANs~\cite{tewari2020stylerig,deng2020disentangled,ghosh2020gif,shi2021lifting} for controllable synthesis.
However, most of them are learned for noise-to-image random face generation, 
which does not naturally fit with the image manipulation task.
Hence, they require time-consuming optimization in the test time, 
and the inverted latent codes may not lie in the manifold for high-quality editing~\cite{xia2021gan, tov2021designing}.
Moreover, photo-realistic synthesis remains challenging~\cite{kowalski2020config,shi2021lifting}.

%P6: The representative performance of our framework and the comparison with prior art.
To this end, we propose \textbf{\Ours}, a novel framework particularly designed for high-quality \textbf{3D}-controllable \textbf{F}ace \textbf{M}anipulation.
Specifically, we perform a learning process to solve the image-to-image translation/editing problem with a conditional StyleGAN~\cite{karras2020analyzing}.
Different from prior 3D GANs trained for random sampling, 
we train our model exactly for existing face manipulation and do not require optimization/manual tuning after the learning phase.
As shown in Fig.~\ref{fig:teaser},
with a single input face image, our framework manages to produce photo-realistic disentangled editing on attributes of head pose, facial expression, and scene illumination, 
while faithfully preserving the face identity.

Our framework leverages face reconstruction networks and a physically-based renderer, 
where the former estimate the input 3D coefficients and the latter embeds the desired manipulations, e.g., pose rotation, into an identity-preserved rendered face. 
A StyleGAN~\cite{karras2020analyzing} conditional generator then takes in both the original image and the manipulated face rendering to synthesize the edited face.
The consistent identity information provided by the input and the rendered edit signals 
spontaneously creates a strong synergy for identity preservation in manipulation.
Moreover, 
we develop two essential training strategies, reconstruction and disentangled training, 
to help our model gain abilities of identity preservation and 3D editability.
As we find an interesting trade-off between identity and editability in the network structure and the simple encoding strategy is sub-optimal,
we propose a novel multiplicative co-modulation architecture for our framework.
This structure stems from a comprehensive study to understand how to encode different information in the generator's latent spaces,
where it achieves the best performance.
We conduct extensive qualitative and quantitative evaluations on our model and demonstrate good disentangled editing ability, strong identity preservation, and high photo-realism, 
outperforming the prior arts in various tasks.
More interestingly, our model can manipulate artistic faces which are out of our training domain, indicating its strong generalization ability.

%P7: Contribution.

Our contributions can be summarized as follows. 

(1) We propose \Ours, 
a novel conditional GAN framework that is specifically designed for precise, explicit, high-quality, 3D-controllable face manipulation.
Unlike prior works, our training objective is strongly consistent with the task of existing face editing, 
and our model does not require any optimization/manual tuning after the end-to-end learning process. 

(2) We develop two essential training strategies, 
reconstruction and disentangled training to effectively learn our model. 
We also conduct a comprehensive study of StyleGAN's latent spaces for structural design, 
leading to a novel multiplicative co-modulation architecture with strong identity-editability trade-off.

(3) Extensive quantitative
and qualitative evaluations demonstrate the advantage of our method over prior arts.
Moreover, our model also shows a strong generalizability to edit artistic faces, which are out of the training domain.

% {\color{red} which brings photo-realism from GANs 
% and explainability in 3DMMs together to achieve controllable and high-quality face synthesis 
% in the manner of conditional GANs.}

% Our model is learned with pairs of photos and rendered faces, 
% which do not require any human annotations.
% More specifically, we form our training dataset with FFHQ~\cite{karras2019style}
% and a synthesizing pipeline~\cite{deng2020disentangled} that can generate multiple images with the same identity.

\section{Related Works}

\noindent\textbf{3D Face Modelling.} 
3D morphable models (3DMMs)~\cite{parke1974parametric,blanz1999morphable,paysan20093d,li2017learning,booth2018large,tran2019towards} have long been used for face modelling. 
In 3DMMs, human faces are normally parame-trized by texture, shape, expression, skin reflectance and scene illumination in a disentangled manner to enable 3D-controllable face synthesis.
However, 3DMMs require expensive data of 3D~\cite{paysan20093d} or even 4D~\cite{li2017learning} scans of human heads to build, and the rendered images often lack photo-realism due to the low-dimensional linear representation as well as the absence of modelling in hair, mouth cavity, and fine details like wrinkles.
With 3DMMs, many methods attempt to estimate the 3D parameters of  2D images~\cite{ververas2020slidergan,slossberg2018high,sengupta2018sfsnet,sela2017unrestricted,saito2017photorealistic,richardson2017learning,genova2018unsupervised,gecer2019ganfit,sanyal2019learning,deng2019accurate}.
This is normally achieved by optimization or a neural networks to extract 3D parameters from a face image and/or landmarks~\cite{sengupta2018sfsnet,genova2018unsupervised,deng2019accurate}.
In our work, we use 3DMMs for 3D face representation and adopt face reconstruction networks to provide the basis of 3D editing signals from 2D images.  
As such, our model can be trained solely with 2D images to gain 3D controllability.

\noindent\textbf{GAN.} 
Recently, unconditional GANs show promising results in synthesizing photo-realistic faces~\cite{karras2017progressive,karras2019style,karras2020analyzing}.
While latent space exploration~\cite{harkonen2020ganspace,shoshan2021gan,shen2021closed} has proved to be effective,
it requires extensive human labors to obtain meaningful control for generation.
As such, a rich set of literatures propose to use conditional GANs~\cite{bao2018towards,choi2018stargan,pumarola2018ganimation,tran2017disentangled,huang2017beyond,usman2019puppetgan,shen2018faceid,xiao2018elegant,qian2019unsupervised,zhou2019deep} for controllable identity-preserved image synthesis by disentangling identity and non-identity factors.

Noticeably, DR-GAN~\cite{tran2017disentangled} and TP-GAN~\cite{huang2017beyond} disentangles identity and pose to allow frontal view synthesis,
while Zhou et al.~\cite{zhou2019deep} extracts spherical harmonic lighting from source image for portrait relighting.
However, these works can only manipulate one attribute of the faces,
whilst we are able to conduct disentangled editing on pose, lighting, and expression under a unified framework.
We even outperform some of them on the tasks that they are solely trained on.

To transfer multiple attributes,  
Bao et al.~\cite{bao2018towards} and Xiao et al.~\cite{xiao2018elegant} extract identity from one image and facial attributes from another one for reference-based generation.
StarGAN~\cite{choi2018stargan} leverages multiple labeled datasets to learn attributes translation.
In contrast, our model provide manipulations with just a single input, and it does not require any labeled information/datasets for training.

\noindent\textbf{3D Controllable GAN.}
In line with our work, several prior methods~\cite{nguyen2019hologan,deng2020disentangled,kowalski2020config,tewari2020stylerig,tewari2020pie,ghosh2020gif,shi2021lifting,buehler2021varitex}
introduce 3D priors into GANs to achieve 3D controllability over face attributes of expression, pose, and illumination.

Deng et al.~\cite{deng2020disentangled} enforce the input space of its GAN to bear the same disentanglement as 
the parameter space of a 3DMM  to achieve controllable face generation.
GIF~\cite{ghosh2020gif} conditions the space of StyleGAN's layer noise on render images from FLAME~\cite{li2017learning} to control pose, expression, and lighting.
Tewari et al.~\cite{tewari2020stylerig} leverage a pretrained StyleGAN and learn a RigNet to manipulate latent vectors with respect to the target editing semantics. 
However, these approaches are all trained for random face generation, not for existing face manipulation. 
Although GAN inversion can well project existing images in their latent spaces for good reconstruction, 
these latent codes may not fall on the manifold with good editability, 
leading to noticeable quality drop after manipulation.
On the contrary, our model is trained exactly for the task of real face editing, 
which demonstrates a clear improvement in manipulation quality upon these works.

While CONFIG~\cite{kowalski2020config} does not need GAN inversion for real image manipulation,
its parametric editing space doesn't inherit identity information from the input images, 
resulting in a clear identity loss.
Moreover, our novel generator architecture also provides us with a larger range of editability and higher photo-realism upon them.
While PIE~\cite{tewari2020pie} proposes a specialized GAN inversion process to be later combined with StyleRig for real image manipulation,
we again find our approach provides better quality and higher efficiency.
Compared to a more recent VariTex~\cite{buehler2021varitex} approach which can not synthesize background and rigid body like glasses, 
our method produces much more realistic outputs.

% Our work is also related to StylePoseGAN~\cite{sarkar2021style} which uses StyleGAN to synthesize novel views of human body.
% However, while they just study a naive exclusive modulation scheme to encode the edit and identity signals, 
% we conduct a more comprehensive study to understand which signals should go to what latent spaces of a StyleGAN to improve the identity-editability tradeoff.

\section{Methodology}

\subsection{Overview and Notations}

\begin{figure}[t]
    \centering
    \includegraphics[width = 0.96\textwidth]{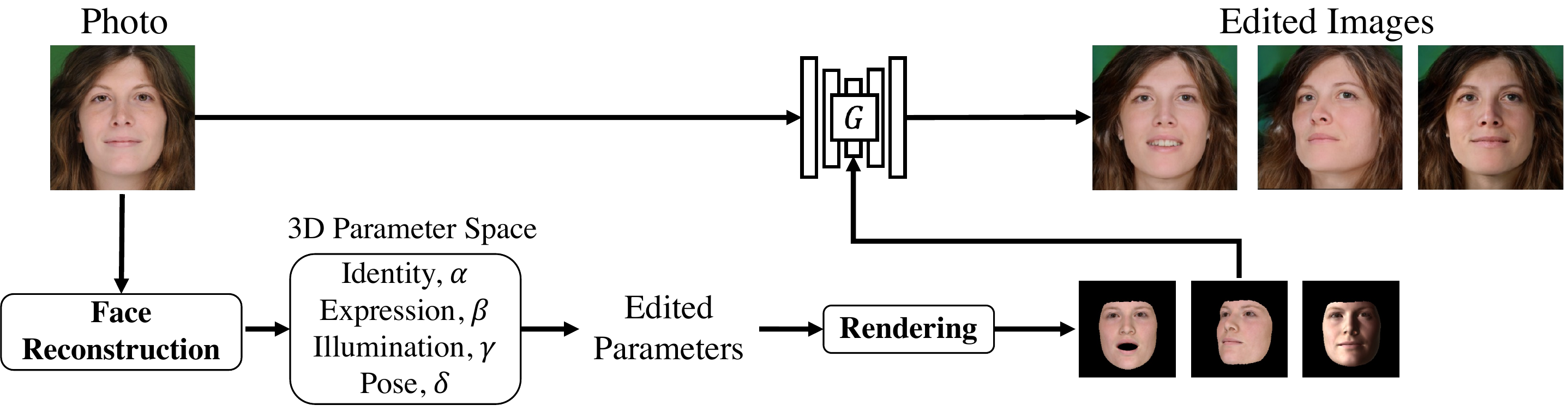}
    \vspace{-0.35cm}
    \caption{Workflow of \Ours. 
    Given a photo input, our framework first extracts its 3D parameter by face reconstruction and then renders identity-preserved 3D-manipulated edit faces.
    The input photo and the edit signals are later jointly encoded into a generator to synthesize various photo editings.
    } %for 3D-controllable identity preserved image synthesis.}
    \label{fig:framework}
    \vspace{-0.6cm}
\end{figure}

In Fig.~\ref{fig:framework}, we show the workflow of \Ours\, 
which consists of:
the generator $\G$, the face reconstruction network $\FR$, and the renderer $\Rd$. 
Given an input face image $P \in \mathbb{R}^{H\times W\times 3}$,
it first estimates the lighting and 3DMM parameters of the face $p~\text{=}~\FR(P)$, $p~\text{=}~(\alpha, \beta, \gamma, \delta) \in \mathbb{R}^{254}$. Naturally, $p$ has disentangled controllable components for identity $\alpha \in \mathbb{R}^{160}$, expression $\beta \in \mathbb{R}^{64}$, lighting $\gamma \in \mathbb{R}^{27}$, and pose $\delta \in \mathbb{R}^{3}$~\cite{paysan20093d}.
The disentangled editing is then achieved in this parameter space, 
where we keep the identity factor $\alpha$ unchanged but adjust $\beta, \gamma, \delta$ to the desired semantics for the expression, lighting, and pose, 
which returns a manipulated parameter $\hat{p}$ to render an image $\hat{R}~\text{=}~\Rd(\hat{p})$~\cite{blanz1999morphable}.
Finally, the manipulated photo is generated by feeding $P$ and $\hat{R}$ through the generator as $\hat{P}~\text{=}~\G(P, \hat{R})$.
In this way, the synthesized output $\hat{P}$ will preserve the identity from $P$,
while its expression, illumination, and pose, follow the control from $\hat{p}$.

\subsection{Dataset}

\begin{wrapfigure}{r}{0.5\textwidth}
    \vspace{-0.7cm}
    \centering
    \includegraphics[width=0.5\textwidth]{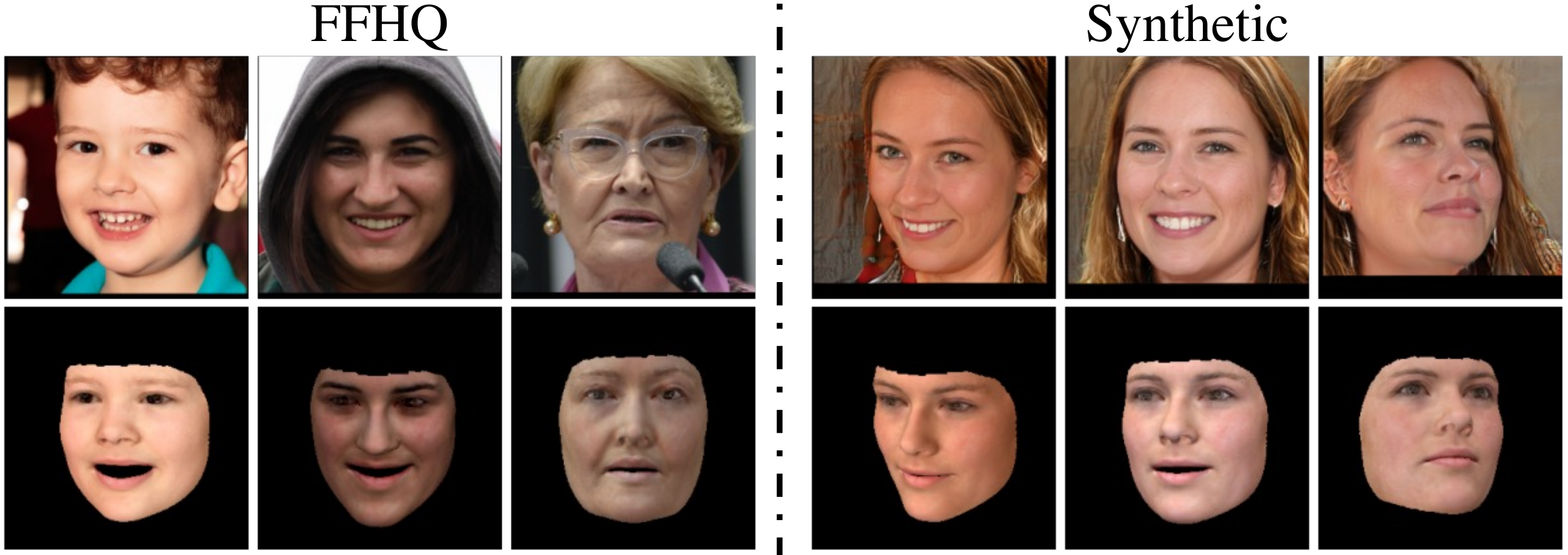}
    \vspace{-0.7cm}
    \caption{Examples of photo and render image pairs ($P$, $R$). 
    \textbf{Left:} FFHQ data. Each identity just has one corresponding image. 
    \textbf{Right:} Synthetic data from~\cite{deng2020disentangled}. We generate multiple images for an identity with varied expression, pose, and illumination.}
    \label{fig:dataset}
    \vspace{-0.8cm}
\end{wrapfigure}

The training data are in the form of photo and render image pairs  ($P$, $R$), where $P$ and $R$ share the same attributes of identity, expression, illumination, and pose.
We construct our dataset with both the FFHQ data and synthetic data. We show examples of the data pairs in Fig.~\ref{fig:dataset}.

\noindent\textbf{FFHQ.}
FFHQ~\cite{karras2019style} is a human face photo dataset,
where most identities only have one corresponding image.
For each of the training image $P$, we extract its render counterpart by $R~ \text{=}~\Rd(\FR(P))$ to form the ($P$, $R$) pair.

\noindent\textbf{Synthetic Dataset.}
We also require a dataset where each identity has multiple images with various attributes of expression, pose, and illumination.
Such a dataset is crucial for model to perform learning for editing.
While this kind of high-quality dataset is not publicly available, 
we leverage DiscoFaceGAN~\cite{deng2020disentangled}, $\Gd$, to synthesize one as follows.

Given a parameter $p$ of our 3D parameter space and a noise vector $n \in \mathbb{R}^{32}$, 
$\Gd$ synthesizes a photo image $P~\text{=}~\Gd(p, n)$ that resembles the identity, expression, illumination, and pose of its render counterpart $R~\text{=}~\Rd(p)$.
We can thus generate multiple images of the same identity with other attributes varied following the steps below: 
(1) randomly sample a 3D parameter $p_1~\text{=}~(\alpha_1, \beta_1, \gamma_1, \delta_1)$ and a noise $n$;
(2) keep $\alpha_1$ unchanged and re-sample $M~\text{-}~1$ tuples of $(\beta, \gamma, \delta)$ such that we have $p_2$ = ($\alpha_1$, $\beta_2$, $\gamma_2$, $\delta_2$), ..., $p_M$ = ($\alpha_1$, $\beta_M$, $\gamma_M$, $\delta_M$);
(3) Use $\Gd$ and $\Rd$ to generate photo-render pairs of ($P_1$, $R_1$), ... ($P_M$, $R_M$), where $P_i$ = $\Gd(p_i, n)$ and $R_i~\text{=}~\Rd(p_i)$.
Such process is iterated for $N$ identities to form a dataset of $N \times M$ pairs.
Examples of such image pairs are in Fig.~\ref{fig:dataset} (\textbf{Right}).

\subsection{Training Strategy}\label{sec:train_strat}

\begin{figure}[t]
    \centering
    \includegraphics[width=\textwidth]{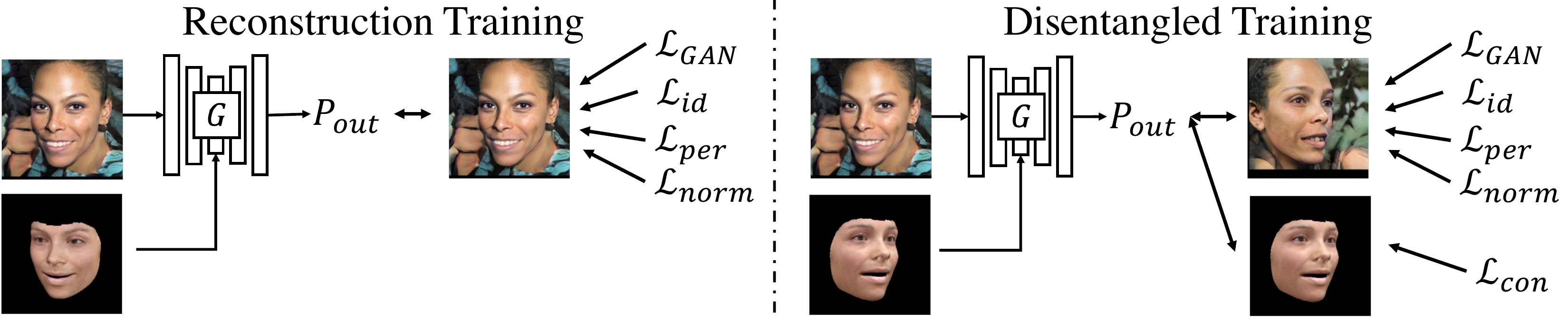}
    \vspace{-0.6cm}
    \caption{Proposed model learning strategies of reconstruction training (\textbf{Left}) and disentangled training (\textbf{Right}).}
    \label{fig:training}
    \vspace{-0.5cm}
\end{figure}

While $\FR$ and $\Rd$ do not require further tuning,  
we design two strategies, 
reconstruction and disentangled training (Fig.~\ref{fig:training}) to train $\G$.
We find the former helps for identity preservation while the latter ensures editability (Fig.~\ref{fig:ablation_study}). 
Formally, we denote the input pair $(P_{in}, R_{in})$ and its output $P_{out} = \G(P_{in}, R_{in})$.

\noindent\textbf{Reconstruction Training.}
We first equip $\G$ with the ability to reconstruct $P_{in}$ from ($P_{in}$, $R_{in}$).
In this case, we want $P_{out}$ to be as similar as $P_{in}$, and set the target output $P_{tg}~\text{=}~P_{in}$.
We first define a face identity loss with a face recognition network $\Nf$~\cite{deng2019arcface} :
\vspace{-0.25cm}
\begin{equation}\label{eqn:id_loss}
\vspace{-0.25cm}
\L_{id} = ||\Nf(P_{out}) - \Nf(P_{tg})||_2^2
\end{equation}
We also enforce $P_{out}$ and $P_{tg}$ to have similar low-level features and high-level perception by imposing an $\ell$1 loss and a perceptual loss based on LPIPS~\cite{zhang2018unreasonable}:
\vspace{-0.25cm}
\begin{equation}\label{eqn:l1_loss}
\vspace{-0.25cm}
\L_{norm} = ||P_{out} - P_{tg}||_1
\end{equation}
\vspace{-0.25cm}
\begin{equation}\label{eqn:per_loss}
\vspace{-0.15cm}
\L_{per} = LPIPS(P_{out}, P_{tg})
\end{equation}
Finally, we adopt the GAN loss, $\L_{GAN}$ such that the generated images $P_{out}$ shall match the distribution of $P_{tg}$.
In this way, our loss is constructed as:
\vspace{-0.25cm}
\begin{equation}
\vspace{-0.25cm}
\L_{rec} = \L_{GAN} + \lambda_1\L_{id} + \lambda_2\L_{norm} + \lambda_3\L_{per}
\end{equation}
where $\lambda_1, \lambda_2, \lambda_3$ are the weights for different losses.
We use both synthetic and FFHQ datasets for this procedure.

\noindent\textbf{Disentangled Training.}
To achieve our goal, 
only teaching the model how to ``reconstruct'' is not sufficient. Thus, we propose a disentangled training strategy to enable editing, 
which can only be achieved by the synthetic dataset as it has multiple images of the same identity with varying attributes.

Specifically, we first sample two pairs,  ($P_{in}^1$, $R_{in}^1$) and ($P_{in}^2$, $R_{in}^2$), from the same identity.
Then, given $P_{in}^1$ and $R_{in}^2$, we want our model to produce $P_{in}^2$, which shares the same edit signal in $R_{in}^2$ while the same identity of $P_{in}^1$.
In this case, we set $P_{out}^1 = \G(P_{in}^1, R_{in}^2)$ and $P_{tg}^1 = P_{in}^2$,
and impose the prior defined loss of $\L_{GAN}$, $\L_{id}$, $\L_{norm}$, and $\L_{per}$ between $P_{out}^1$ and $P_{tg}^1$.
Different from reconstruction, 
we also inject a content loss to better capture the target editing signals, where we set $R_{tg}^1 = R_{in}^2$ and define the loss as:
\vspace{-0.35cm}
\begin{equation}\label{eqn:con_loss}
\vspace{-0.15cm}
\L_{con} = ||M\odot (P_{out}^1 - R_{tg}^1) ||_2^2
\end{equation}
$M$ is the face region that $R^1_{tg}$ has non-zero pixels and $\odot$ is the element-wise multiplication.
To sum up, the loss of our disentangled training is:
\vspace{-0.25cm}
\begin{equation}
\vspace{-0.25cm}
\L_{dis} = \L_{GAN} + \lambda_1\L_{id} + \lambda_2\L_{norm} + \lambda_3\L_{per} + \lambda_4\L_{con}
\end{equation}
with the same weights of $\lambda_1, \lambda_2, \lambda_3$ as reconstruction training.
To use the loaded data more efficiently, we repeat the same procedure for $P_{out}^2 = \G(P_{in}^2, R_{in}^1)$.

\noindent\textbf{Learning Schedule.}
In practice, we alternate between reconstruction and disentangled training: for every $S$ iterations, 
we do 1 step of disentangled training and $S$ - 1 steps of reconstruction.
Moreover,
as reconstruction can be performed by both synthetic and FFHQ datasets, 
we carry out our learning in two phases.
In phase-1, we takes synthetic data for both training strategies. 
In phase-2, we switch to FFHQ for reconstruction while still uses synthetic data for disentangled training.
Fig.~\ref{fig:ablation_study} shows the advantages of this 2-phase learning.

\subsection{Architecture}

Our conditional generator $\G$ is composed of a set of encoders $\E$ and a StyleGAN~\cite{karras2020analyzing} generator $\mathbf{G_s}$.
We utilize three latent spaces of $\mathbf{G_s}$ for information encoding, namely, the input tensor space $\T \in \Real^{512 \times 4 \times 4}$, the modulation space $\W \in \Real^{512}$, and the extended modulation space $\W^{+} \in \Real^{512 \times L}$ ($L$: number of layers in $\mathbf{G_s}$).
We denote the encoder to each of these spaces as $\mathbf{E_T}$, $\mathbf{E_W}$, and $\mathbf{E_{W^{+}}}$ 
and we conduct the study of what ``information" (photo $P$ or render $R$) to be encoded into which ``space" ($\T$, $\W$, and $\W^{+}$),
where we experiment both exclusive modulation and co-modulation architectures.

\begin{figure}[t]
    \centering
    \includegraphics[width = \textwidth]{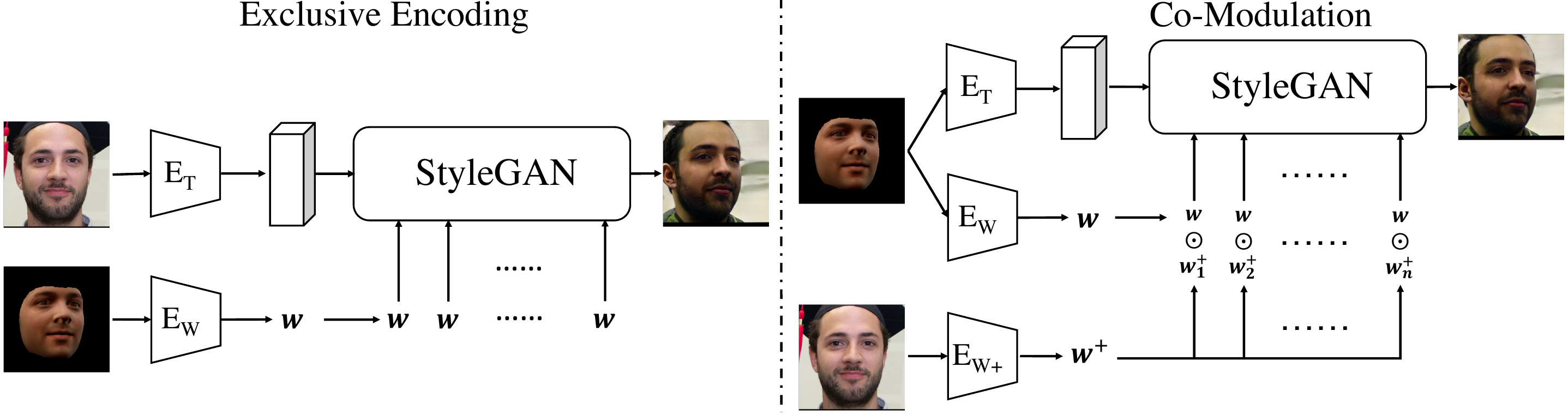}
    \vspace{-0.5cm}
    \caption{Our generator $\G$ can take forms of both exclusive modulation (\textbf{Left}) and co-modulation (\textbf{Right}) architectures.}
    \label{fig:architecture}
    \vspace{-0.5cm}
\end{figure}

\noindent\textbf{Exclusive Modulation.}
Naively, we can exclusively encode $P$ or $R$ into the modulation space  ($\W$/$\W^{+}$).
While one is encoded into the modulation space, 
the other one can only be encoded into $\T$.
Fig.~\ref{fig:architecture} (\textbf{Left}) shows an example of such architectures, 
where $R$ is encoded into $\W$ and $P$ is encoded into $\T$.
This structure is denoted as Render-$\W$.
Whether $R$ or $P$ is used for modulation and whether the space is $\W$ or $\W^{+}$ provides us with 4 variants of exclusive modulation in total, 
and we investigate all of them.

\noindent\textbf{Co-Modulation.}
We further investigate to encode both $P$ and $R$ into $\W$ and $\W^{+}$ 
and combine their embeddings for final modulation.
A representative of such a architecture is shown in Fig.~\ref{fig:architecture} (\textbf{Right}), 
where $R$ is encoded into $\W$, and $P$ is encoded into $\W^{+}$.
In particular, the modulation signal for layer $l$ is obtained by  $\W^{+}_{l}\odot\W$ where $\W^{+}_{l} \in \Real^{512}$ is the $l$-th column of $\W^{+}$ and $\odot$ is the element-wise multiplication.
Unlike prior works that use concatenation or a recent approach of tensor transform plus concatenation scheme~\cite{zhao2021large} to combine $\W^{+}_l$ and $\W$, 
we find our multiplicative co-modulation owns the best effectiveness.
Moreover, in Fig.~\ref{fig:architecture}, we also encode $R$ into $\T$ to further improve the identity-editability trade-off.

\section{Experiment}

\subsection{Experimental Setup}

We adopt the face reconstruction network $\FR$~\cite{deng2019accurate}, the 3DMM~\cite{paysan20093d}, and the renderer $\Rd$~\cite{blanz1999morphable}.
Our conditional generator $\G$ consists of 
the StyleGAN~\cite{karras2020analyzing} generator $\mathbf{G_s}$ and ResNet~\cite{he2016deep} encoders $\E$.
Specifically, $\mathbf{E_T}$ and $\mathbf{E_W}$ are based on ResNet-18 structure, 
where $\mathbf{E_T}$ outputs the feature prior to final pooling and $\mathbf{E_W}$ outputs the layer after that.
We use a 18-layer PSP encoder~\cite{richardson2021encoding} as $\mathbf{E_{W^+}}$.
The discriminator architecture is the same as~\cite{karras2020analyzing}. 
The synthetic data are generated by DiscoFaceGAN~\cite{deng2020disentangled}, 
where we set $N$ and $M$ to 10000 and 7. 
We use the first 65k FFHQ images (sorted by file names) for training 
and the rest 5k images as held-out testing set.
All images (render and photo input, model output) are of 256px resolution.
We set $S$ to 2 and use a batch size of 16 for both reconstruction and disentangled training.
We set $\lambda_1$, $\lambda_2$, $\lambda_3$, and $\lambda_4$ to be 3, 3, 30, and 20.
The model is learned in 2-phase, 
where phase-1 takes 140k iterations followed by 280k updates of phase-2.

\subsection{Evaluation Metrics}\label{sec:eval_metrics}

We develop several quantitative metrics with the held-out 5k FFHQ images (denoted as $\P$) to evaluate identity preservation (\textbf{Identity}), editing controllability (\textbf{Face Content Similarity and Landmark Similarity}), and photo-realism (\textbf{FID}) of our model $\G$ for image manipulation.

\noindent\textbf{Manipulated Images.}
For each $P \in \mathbb{P}$, 
we first get $p$ = $\FR(P)$ = $(\alpha, \beta, \gamma, \delta)$ and then re-sample $(\beta, \gamma, \delta)$ to form edited control parameters $\hat{p}$.
The editing signals and the manipulated images are thus  $\hat{R}~\text{=}~\Rd(\hat{p})$ and  $\hat{P}~\text{=}~\G(P, \hat{R})$. 
We generate 4 $\hat{P}$ for each $P$.

\noindent\textbf{Identity.} 
For each ($P$, $\hat{P}$) pair, we measure the identity preservation by computing the cosine similarity of $<\Nf(P), \Nf(\hat{P})>$. 

\noindent\textbf{Landmark Similarity.}
For each ($\hat{P}$, $\hat{R}$) pair,
we use a landmark detection network $\Nl$~\cite{bulat2017far},  
to extracts both of their 68 2D landmarks. 
The similarity metric is defined as $||\Nl(\hat{P}) - \Nl(\hat{R}) ||_2^2$.

\noindent\textbf{Face Content Similarity.}
For each ($\hat{P}$, $\hat{R}$) pair, we follow Eqn.~\ref{eqn:con_loss} to measure the face content similarity. 

\noindent\textbf{FID.}
We denote all edited images as $\hat{\mathbb{P}}$
and measure FID~\cite{heusel2017gans} between $\mathbb{P}$ and $\hat{\mathbb{P}}$ to evaluate $\G$'s photo-realism.

\subsection{Architectures Evaluation}\label{sec:architecture_ablation}

\noindent\textbf{Exclusive Modulation.} 
We first evaluate exclusive modulation architectures in Tab.~\ref{tab:quant_encoding},
where we observe a trade-off between identity preservation and editability.
For example, Photo-$\W^+$ shows the best identity preservation (\textbf{Id}), 
while its editability of landmark (\textbf{LM}) and face content similarity (\textbf{FC}) is the worst.
On the other hand, Photo-$\W$ and Render-$\W^{+}$ owns strong \textbf{LM} and \textbf{FC}, 
yet their \textbf{Id} are much poorer and they even have issues on photo-realism.
Moreover, we see that Render-$\W$ provides us with a decent editability, 
while improves a lot on \textbf{Id} compared to Render-$\W^{+}$ and Photo-$\W$.
For good identity preservation, 
we design a co-modulation architecture based on Render-$\W$ and Photo-$\W^{+}$.

\begin{wraptable}{r}{0.5\textwidth}
    \vspace{-0.7cm}
    \centering
    \fontsize{7}{8}\selectfont
    \begin{tabular}{|c|c|c|c|c|c|}
    \hline
     \multicolumn{2}{|c|}{Model} & \multicolumn{4}{c|}{Metrics} \\  
       \hline
     Mod. Scheme  & Type  & Id.$^{\uparrow}$  & LM.$^{\downarrow}$  & FC.$^{\downarrow}$ & FID$^{\downarrow}$ \\
     \hline
     \multirow{4}{5em}{Exclusive \\ Modulation} & Render-$\W$ & 0.57 & 17.8 & 0.021 & 14.5 \\
     & Render-$\W^+$ & 0.46 & 16.2 & 0.019  & 25.8 \\
     & Photo-$\W$ & 0.50 & 15.6 & 0.018 & 13.7  \\
     & Photo-$\W^+$ & 0.66 & 27.3 & 0.033 & 12.3 \\ 
     \hline
    \multirow{3}{5em}{2-Encoder Co-Mod} 
    & Concat.  & 0.64 & 59.3 & 0.031 & 17.8 \\
    & Tensor  & 0.62 & 24.9 & 0.028 & 18.6 \\
    & Multi.  & 0.66 & 22.2 & 0.025 & 12.4 \\
    
    \hline
    3-E Co-Mod & Multi. & 0.66 & 17.2 & 0.020 & 12.2 \\
    \hline
    
    \end{tabular}
    \vspace{-0.3cm}
    \caption{Quantitative measurement of identity preservation (\textbf{Id}), editing control (\textbf{LM} \& \textbf{FC}), and photo-realism (\textbf{FID}) for different architectures.
    $\uparrow$ means the higher the better, and vice versa for $\downarrow$.}
    \label{tab:quant_encoding}
    \vspace{-0.6cm}
\end{wraptable}

\noindent\textbf{Co-Modulation.}
Based on the study above, we find that encoding $P$ into $\W^{+}$  produces the best identity preservation, 
while encoding $R$ into $\W$ provides good editability.
Thus, we investigate three 2-encoder co-modulation architectures where $\mathbf{E_W}$ encodes $R$ and $\mathbf{E_{W^+}}$ encodes $P$.
Combining $\W^{+}$and $\W$ are achieved via multiplication, concatenation, and a variant of concatenation named tensor transform in~\cite{zhao2021large}.
From Tab.~\ref{tab:quant_encoding}, 
we find the multiplicative co-modulation achieves the best results from all perspectives.
This could be accounted by the fact that modulation itself is a multiplicative operation and thus merging signals together multiplicatively would provide the best synergy.
We further propose a 3-encoder multiplicative co-modulation architecture (bottom of Fig.~\ref{fig:architecture}) to boost the editability, 
which achieves the best trade-off from our observation.

\begin{wrapfigure}{r}{0.5\textwidth}
    \vspace{-0.7cm}
    \centering
    \includegraphics[width=0.5\textwidth]{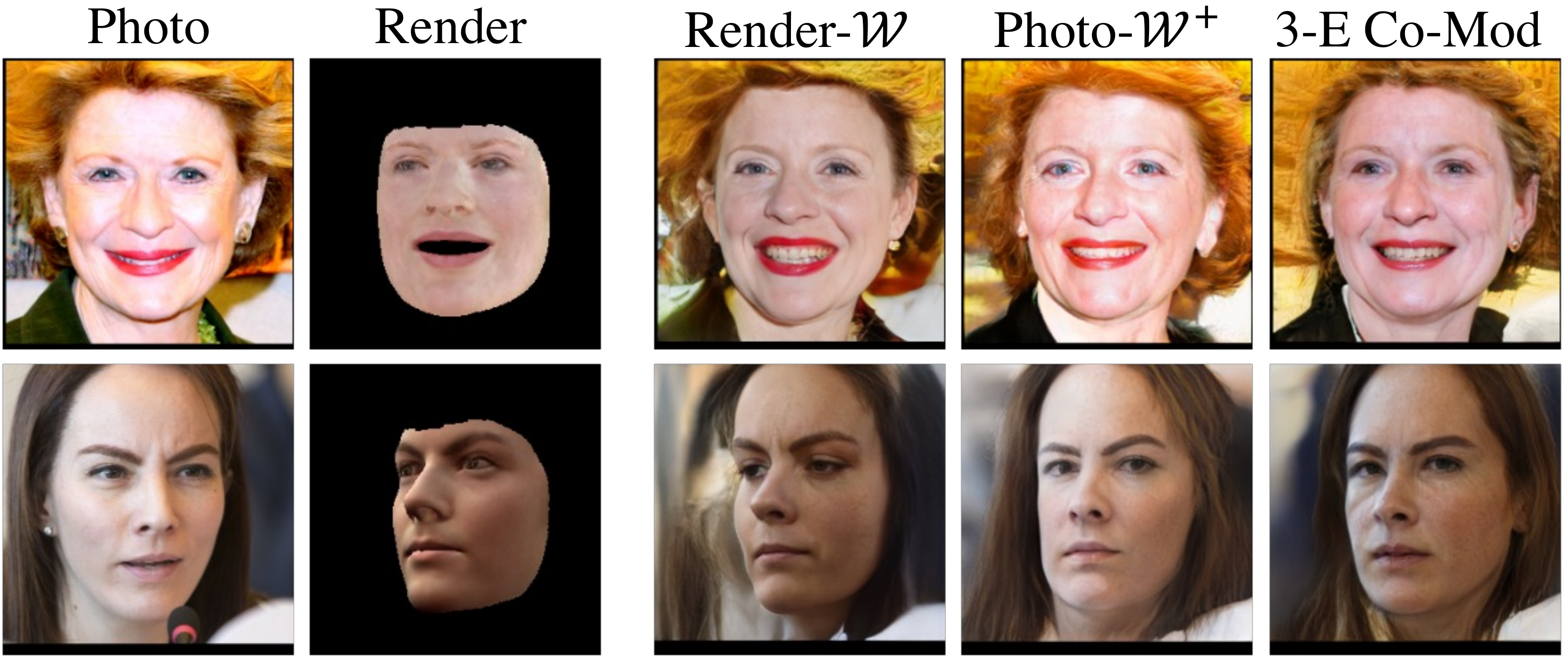}
    \vspace{-0.7cm}
    \caption{Visual comparison among architectures. 
    The co-modulation scheme takes advantages from both:
    good editability from Render-$\W$ and strong identity preservation from Photo-$\W^{+}$.
    }
    \label{fig:archi_ablation_plot}
    \vspace{-0.6cm}
\end{wrapfigure}

\noindent\textbf{Visualization.}
We show a visual comparison among Render-$\W$ (\textbf{Col.~3}), Photo-$\W^{+}$ (\textbf{Col.~4}), and the 3-encoder co-modulation scheme (\textbf{Col.~5}) with the same set of inputs (\textbf{Col.~1 \& 2}) in Fig.~\ref{fig:archi_ablation_plot}.
In the first row, we find that Render-$\W$ has a clear identity loss, while Photo-$\W^{+}$ can hardly manipulate the light intensity, showing inferior editability.
Moreover, Render-$\W$ and Photo-$\W^{+}$ both generate artifacts in the second row.
On the contrary, the co-modulation scheme improves the identity-editability by combining the merits from both schemes: 
good editability from Render-$\W$ and strong identity preservation from Photo-$\W^{+}$.

\subsection{Controllable Image Synthesis}

\begin{figure*}[t]
    \centering
    \includegraphics[width=1.0\textwidth]{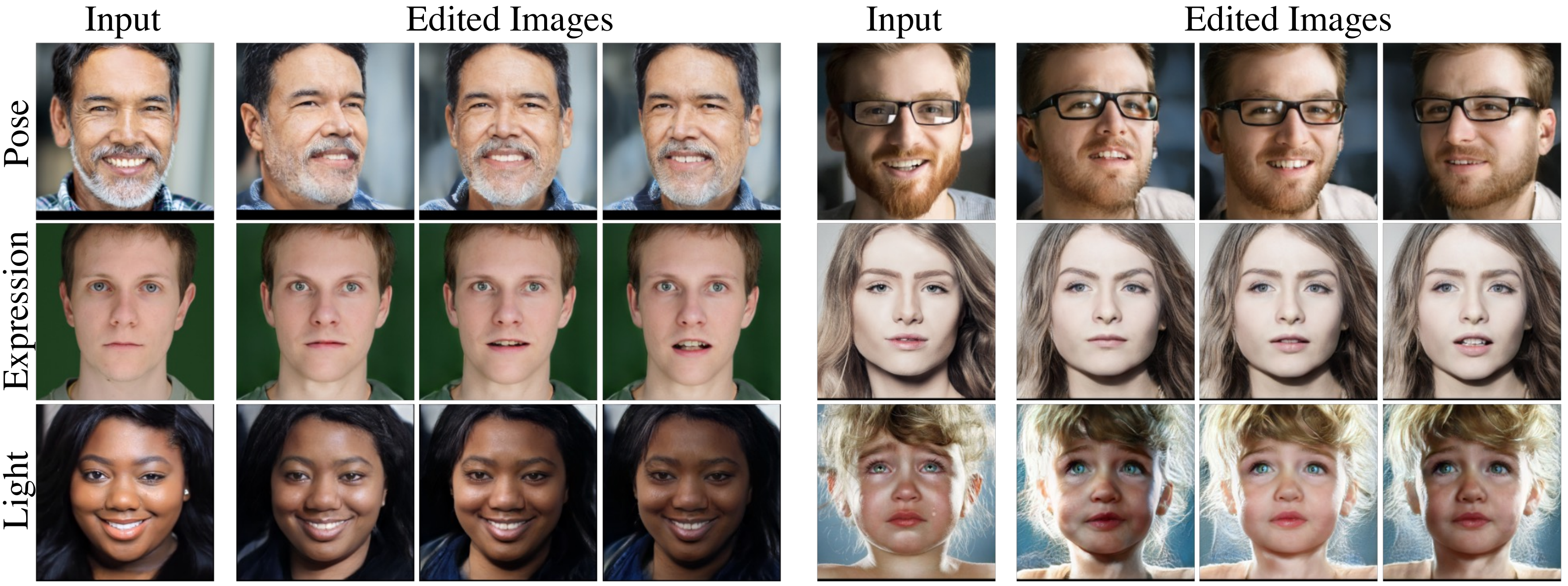}
    \vspace{-0.7cm}
    \caption{Our model provides a variety of disentangled controls for pose (\textbf{Row 1}), expression (\textbf{Row 2}), and illumination (\textbf{Row 3}).
    It shows strong preservation across diverse identities and for facial details like glasses.
    }
    \label{fig:sig_fac_dis_editing}
    \vspace{-0.4cm}
\end{figure*}

We apply our 3-encoder co-modulation architecture to several image manipulation tasks where it all shows good editing controllability, strong identity preservation, and high photo-realism.
More samples are in Supplementary.

\begin{wrapfigure}{r}{0.5\textwidth}
	
    \centering
    \includegraphics[width=0.5\textwidth]{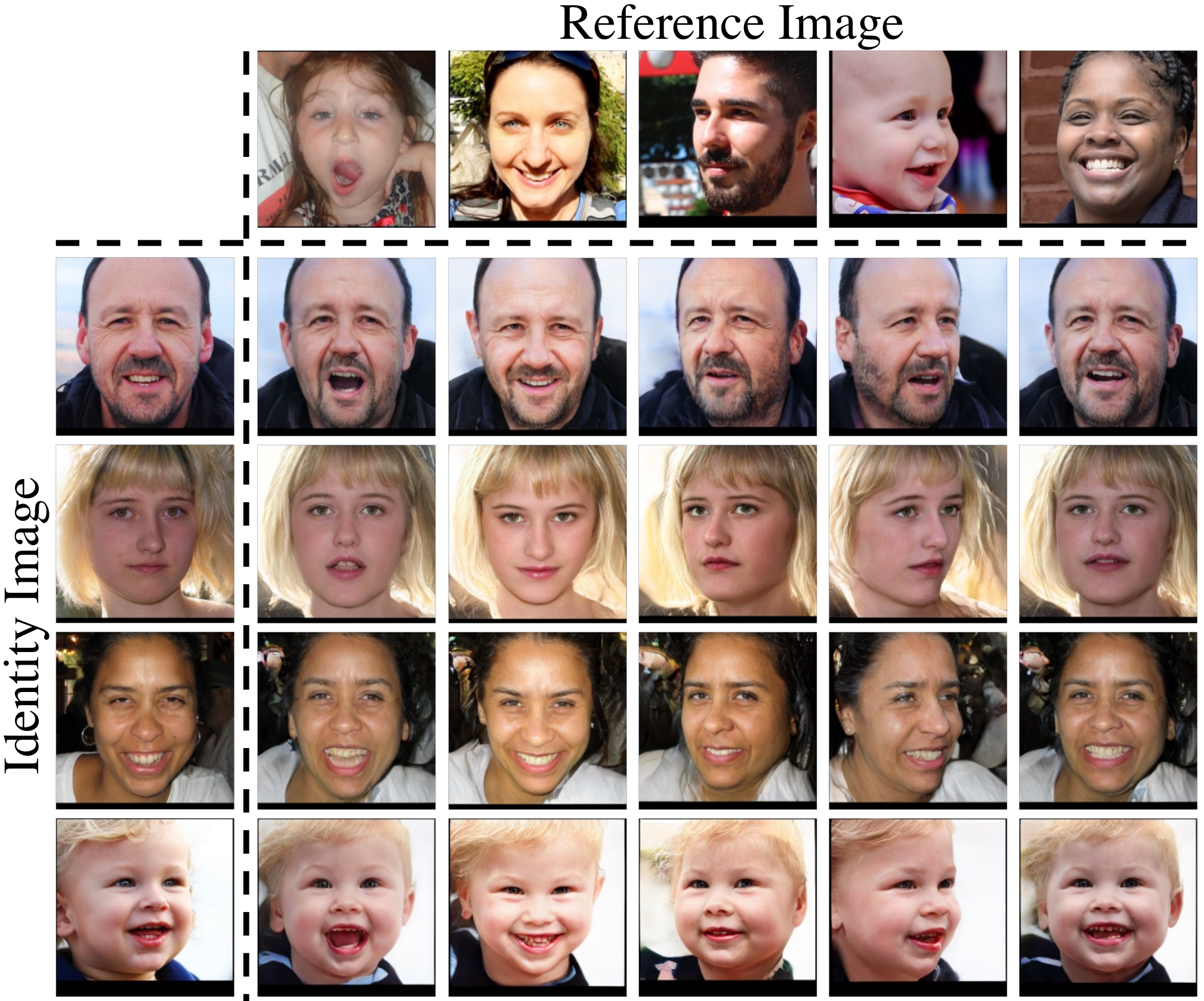}
    \vspace{-0.8cm}
    \caption{Reference based face generation.
    The facial attributes of pose, expression, and illumination are extracted from the reference images to manipulate the identity images.
    }
    \label{fig:ref_face_gen}
    \vspace{0.3cm}
    \includegraphics[width=0.5\textwidth]{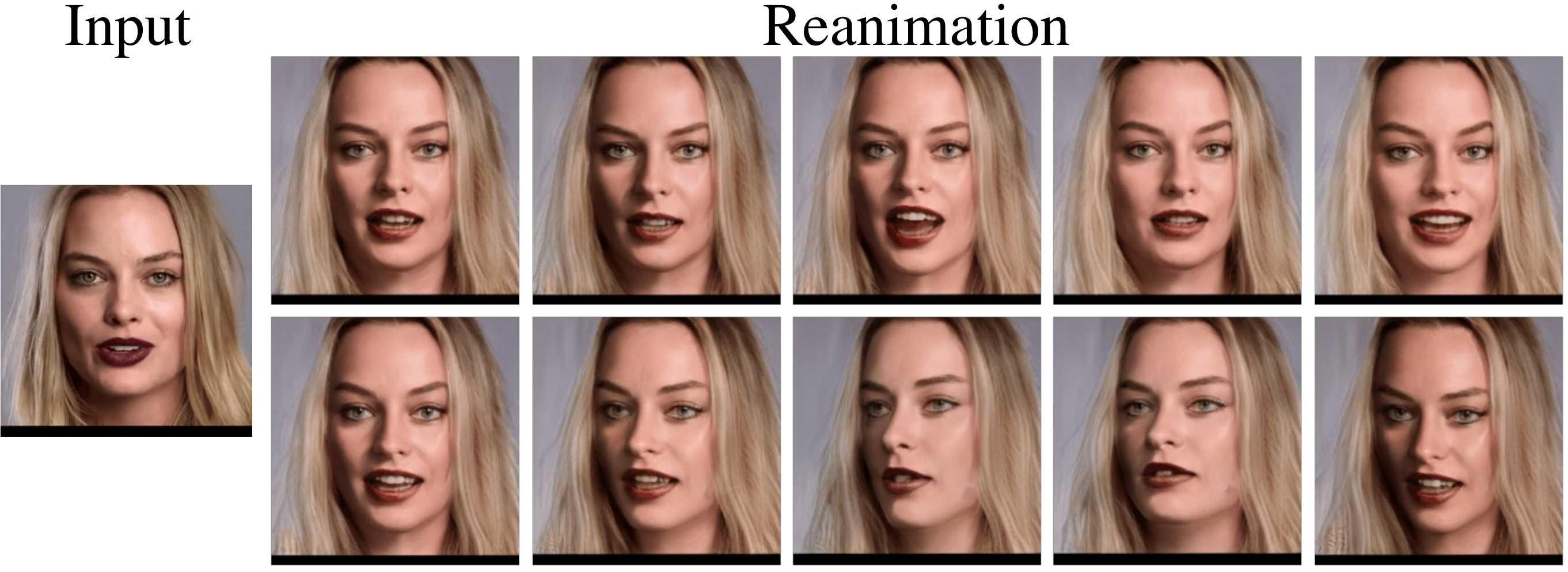}
    \vspace{-0.7cm}
    \caption{Image reanimation. 
    Our model again well preserves the identity and some subtle facial attributes like the dark lipstick. 
    }
    \label{fig:reanimation}
    \vspace{-0.5cm}

\end{wrapfigure}

\noindent\textbf{Disentangled Editing.}
Fig.~\ref{fig:teaser} and Fig.~\ref{fig:sig_fac_dis_editing} show the results of single factor editing,
where we only change one factor of pose, expression, and illumination at a time.
Our model provides highly disentangled editing for the edited factors, while all others remain the same.
Moreover, it shows strong preservation for the identity across people with diverse ages, genders, etc., 
and subtle facial details like the glasses and the teeth.

\noindent\textbf{Reference-Based Synthesis.}
Our model can also perform image manipulation based on reference images shown in Fig.~\ref{fig:ref_face_gen}.
With the pose, expression, and illumination extracted from the reference images, 
we re-synthesize our identity images to bear these editing facial attributes while the identities are still well preserved.

\noindent\textbf{Face Reanimation.}
Our model can also be applied for face reanimation, as shown in Fig.~\ref{fig:reanimation}. 
With a single input photo image, we provide a series of editing render signals to make it animated,
where the identity of the person is well preserved across frames.
Moreover, our model again well preserves facial details like the dark lipstick.

% \begin{figure}[t]
%     \centering
    
% \end{figure}

\begin{wrapfigure}{r}{0.5\textwidth}

    \vspace{-0.7cm}

   \centering    \includegraphics[width=0.5\textwidth]{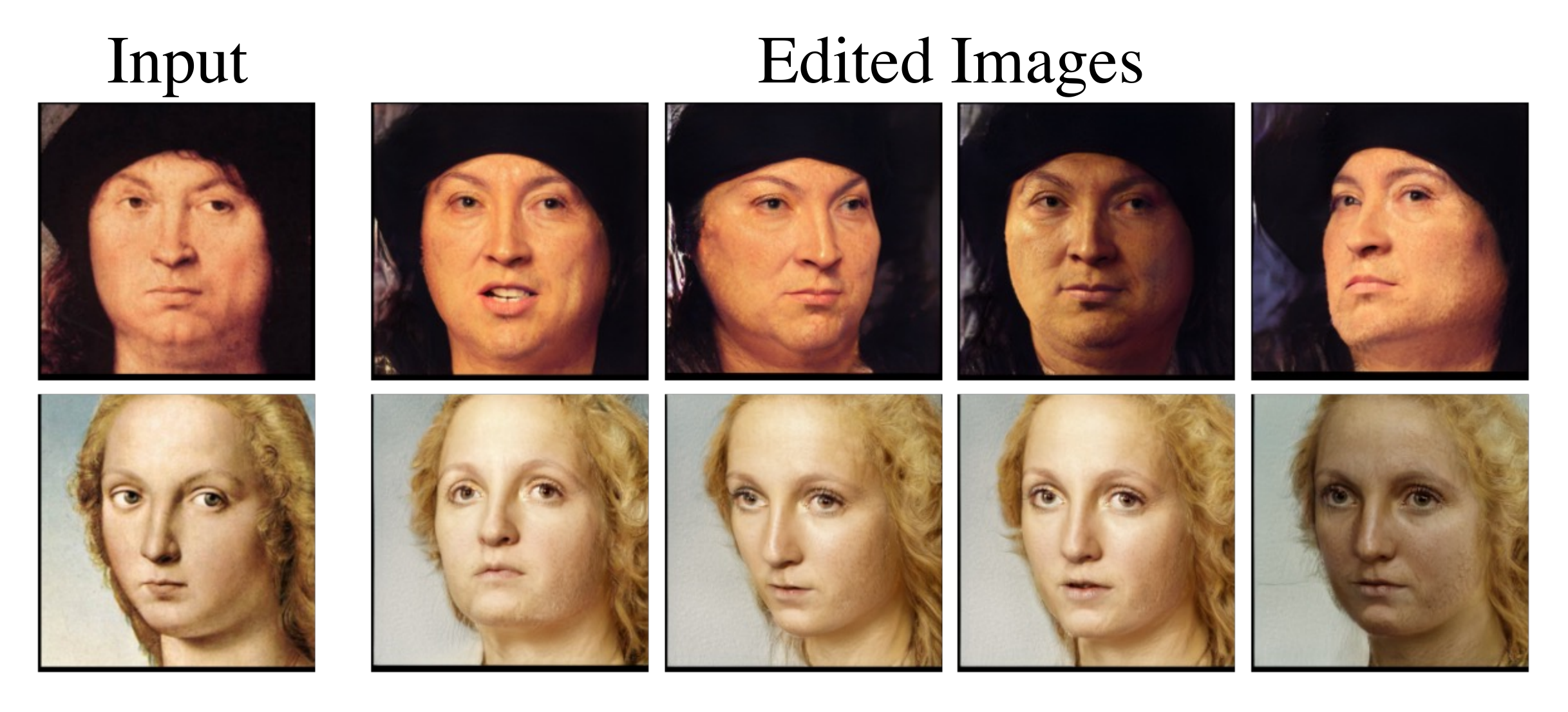}
    \vspace{-0.7cm}
    \caption{Although our model is solely trained on photo faces,
    it demonstrates a strong generalizability to manipulate artistic faces.}
    \label{fig:artistic}
    \vspace{-1.0cm}
\end{wrapfigure}

\noindent\textbf{Artistic Images Manipulation.}
We further perform manipulation on artistic faces~\cite{yaniv2019face} in Fig.~\ref{fig:artistic}.
Surprisingly, 
although our model is only trained on photography faces,
it can still provide controllable and identity-preserved editing on artistic images that are out of the training domain.
This well indicates the strong generalizability of our model. 

% \begin{figure}[t]
%     \centering
    
% \end{figure}

\begin{comment}

    \vspace{-0.7cm}
    \centering

\end{comment}
\section{Comparison to State of the Arts}

We compare with prior 3D-controllable GANs~\cite{deng2020disentangled,tewari2020pie,tewari2020stylerig,shi2021lifting,kowalski2020config,nguyen2019hologan,buehler2021varitex}, 
and show more results in Supplementary.

\begin{figure}[t]
    \centering
    \includegraphics[width=0.9\textwidth]{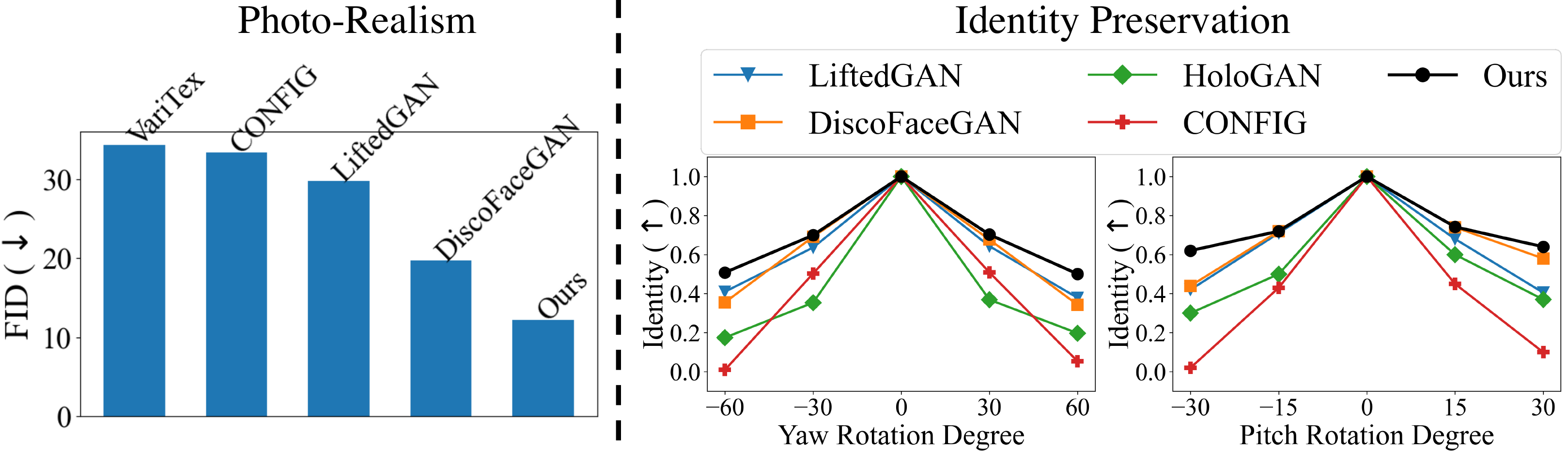}
    \vspace{-0.5cm}
    \caption{Quantitative comparison with prior arts. 
    Our method achieves the best photo realism (\textbf{Left})
    and better identity preservation (\textbf{Right}) at different rotation angles.
    }
    \label{fig:sota_quant_comparison}
    \vspace{-0.4cm}
\end{figure}

\subsection{Quantitative Comparison}

From Fig.~\ref{fig:sota_quant_comparison} (\textbf{Left}), we clearly find that our model produces the most photo-realistic images with the lowest FID. 
We also follow the similar strategy in~\cite{shi2021lifting} to measure identity preservation,
where we use all frontal images from the held-out FFHQ set and perform pose editing at different angles to compute the identity cosine similarity between the edited faces and the original ones.
While prior methods evaluates the preservation between their generated images that naturally fit with their latent manifolds,
we are assessing the identity preservation with real world images, which represents a more challenging task.
Surprisingly, as shown in Fig.~\ref{fig:sota_quant_comparison} (\textbf{Right}),
our model still outperforms prior arts in all rotation angles on a harder task, 
and it can well preserve identity even at large angles.
% While CONFIG~\cite{kowalski2020config} also offers direct embedding of images and we use DiscoFaceGAN~\cite{deng2020disentangled} for data synthesis,
% we improve over them by a clear margin in both identity preservation and photo-realism.

\subsection{Visual Comparison}

\begin{figure}[t]
    \centering
    \includegraphics[width = \textwidth]{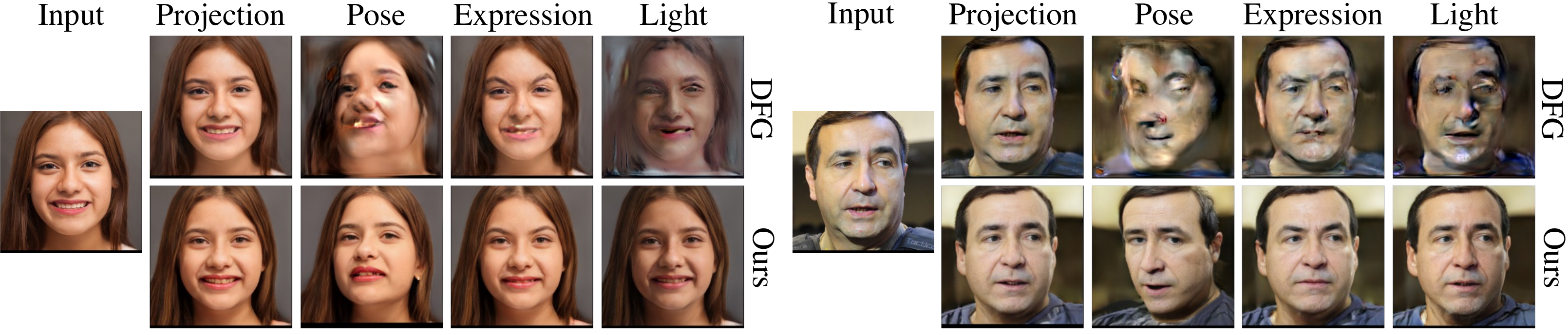}
    \vspace{-0.7cm}
    \caption{Comparing \Ours\ with GAN inversion + DiscoFaceGAN (DFG)~\cite{deng2020disentangled} for face editing.
    Although GAN inversion allows good projection with DFG, providing faithful manipulation with the inverted codes remains challenging.
    On the contrary, our method achieves good reconstruction and high-quality disentangled editing.
    }
    \label{fig:disco_compare_1}
    \vspace{-0.3cm}
\end{figure}

\begin{figure}[t]
    \centering
    \includegraphics[width = 0.85\textwidth]{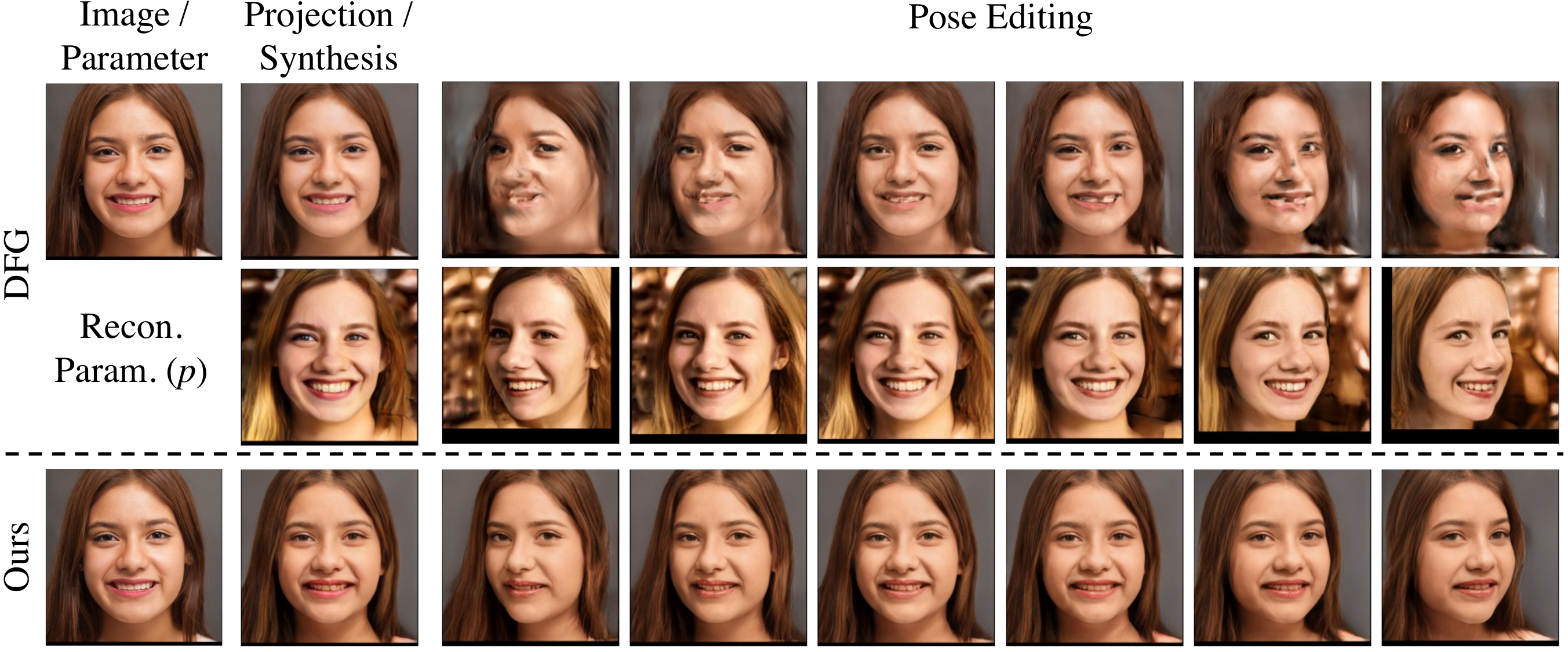}
    \vspace{-0.4cm}
    \caption{Analysis of $\mathcal{W}+$ and $\lambda$ space of DiscoFaceGAN (DFG)~\cite{deng2020disentangled}. 
    While DFG's image embedding is performed in $\mathcal{W}+$ space (\textbf{Row 1}) for editing, 
    we also extract the input's 3DMM parameter  $p$ by $\FR$ and conduct the same editing in $\lambda$ space (\textbf{Row 2}).
    While its $\lambda$ space provides realistic synthesis,
    its inverted code in $\mathcal{W}+$  falls off the manifold of good editability trained in $\lambda$.
    In contrast, \Ours\ (\textbf{Row 3}) uses the same editing spaces for training and testing, 
    which easily leads to high-quality editing.
    }
    \label{fig:disco_compare_2}
    \vspace{-0.4cm}
\end{figure}

\noindent\textbf{DiscoFaceGAN.} 
We first compare \Ours\ with the direct combination of GAN inversion~\cite{abdal2020image2stylegan++} 
+ noise-to-image, 3D GAN, here
DiscoFaceGAN (DFG)~\cite{deng2020disentangled} for image manipulation in Fig.~\ref{fig:disco_compare_1}.
Although GAN inversion successfully retrieves latent codes that well project the image in DFG,
manipulating these codes for high-quality editing is still challenging.
On the contrary, our approach provides both good image reconstruction and high-quality disentangled editing.

We further notice DFG is primarily trained to disentangle its $\lambda$ space where 3DMM parameter $p$ lies in, 
while its image embedding is conducted in $\mathcal{W}+$ space\footnote{In~\cite{deng2020disentangled}, it claims that DFG's $\lambda$ space is not feasible for image embedding.}.
This already creates an obvious disparity between training and test time tasks as different latent spaces are used. 
We thus analyze how DFG behaves in these two spaces in Fig.~\ref{fig:disco_compare_2}, 
where we retrieve $p$ from photo input by  $\FR$ and perform the same editing  in both $\lambda$ and $\mathcal{W}+$ space. 
We clearly see that DFG's $\lambda$ space is well trained for realistic disentangled synthesis, 
yet its $\mathcal{W}+$ space is not.
This suggests that despite the inverted code in $\mathcal{W}+$ can well embed the image, 
it may not lie on the manifold with good editability trained from $\lambda$. 
On the contrary, our method utilizes the same editing space for both training and testing, 
and this consistency guarantees the high-quality manipulation.

\begin{figure}[t]
    \centering
    \includegraphics[width =\textwidth]{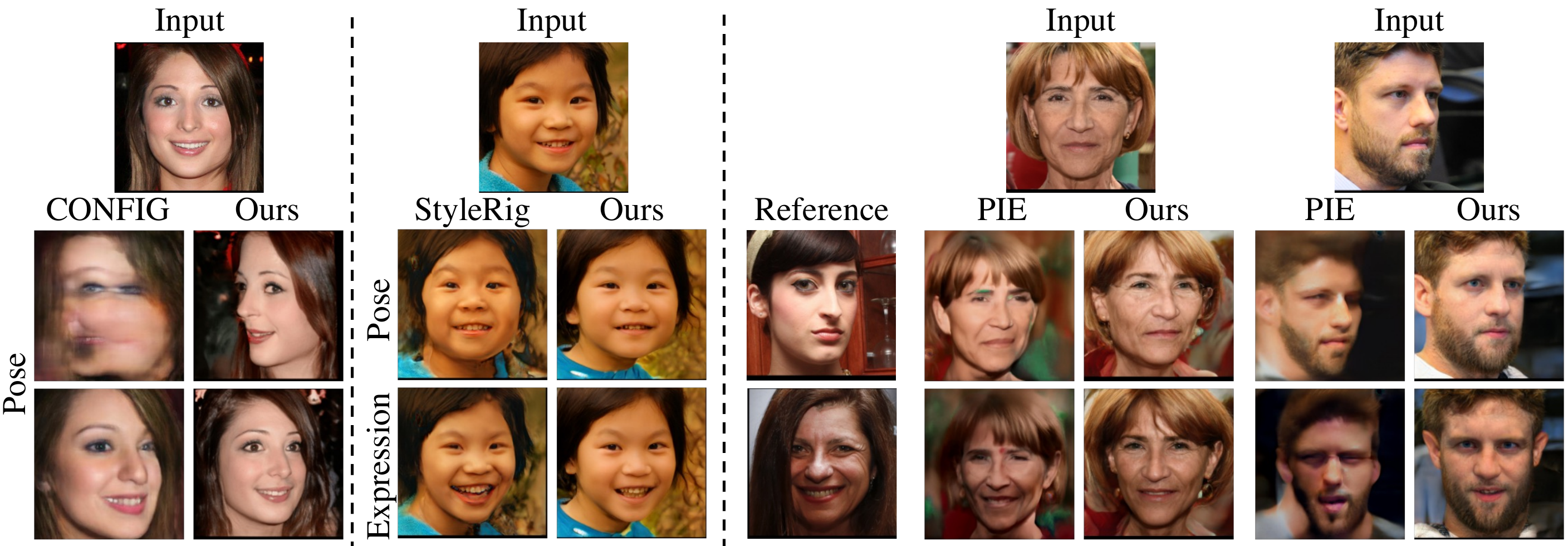}
    \vspace{-0.8cm}
    \caption{Comparing \Ours\ with other 3D-aware GANs for image manipulation.
    Specifically, we compare with CONFIG~\cite{kowalski2020config} on pose editing and StyleRig~\cite{tewari2020stylerig} on both pose and expression editing.
    We compare with PIE~\cite{tewari2020pie} on a reference-based synthesis task, 
    where the pose, expression, and light are extracted from the reference images.
    Our method again shows the best editing results over all prior arts.}
    \label{fig:config_stylerig_pie_comparison}
    \vspace{-0.7cm}
\end{figure}

% Specifically, given a photo $P$, and its extracted 3DMM parameter $p$ = $\FR$($P$), we do the following: 
% (1) $\mathcal{W}+$ space: follow above image manipulation step. 
% (2) $\lambda$ space: sample a noise $n$ and directly synthesize images by feedforward-only inference $\Gd(p, n)$ and $\Gd(\hat{p}, n)$.

\noindent\textbf{Other 3D GANs.}
We further compare \Ours\ with CONFIG~\cite{kowalski2020config}, StyleRig~\cite{tewari2020stylerig}, and PIE~\cite{tewari2020stylerig} on disentangled image editing and reference-based synthesis tasks in Fig.~\ref{fig:config_stylerig_pie_comparison}.
Our model clearly shows a larger range of pose editability and better identity preservation over CONFIG.
Compared to GAN inversion~\cite{abdal2020image2stylegan++} + StyleRig, 
\Ours\ again provides more realistic synthesis with much less artifacts around the face.
While PIE could not provide high-quality manipulation when the input is at large pose rotation angle (\textbf{2nd Example}),
\Ours\ still achieves faithful editing,
indicating its advantage in better generalizability.

% Our method again achieves better manipulation quality compared to prior arts, 
% especially when the input or the target editing pose is at large rotation angle. 
% These again demonstrate the advantage of our approach over concatenation of GAN inversion + 3D-aware, noise-to-image GAN.

\begin{wrapfigure}{r}{0.5\textwidth}
\vspace{-0.7cm}
    \centering
    \includegraphics[width=0.5\textwidth]{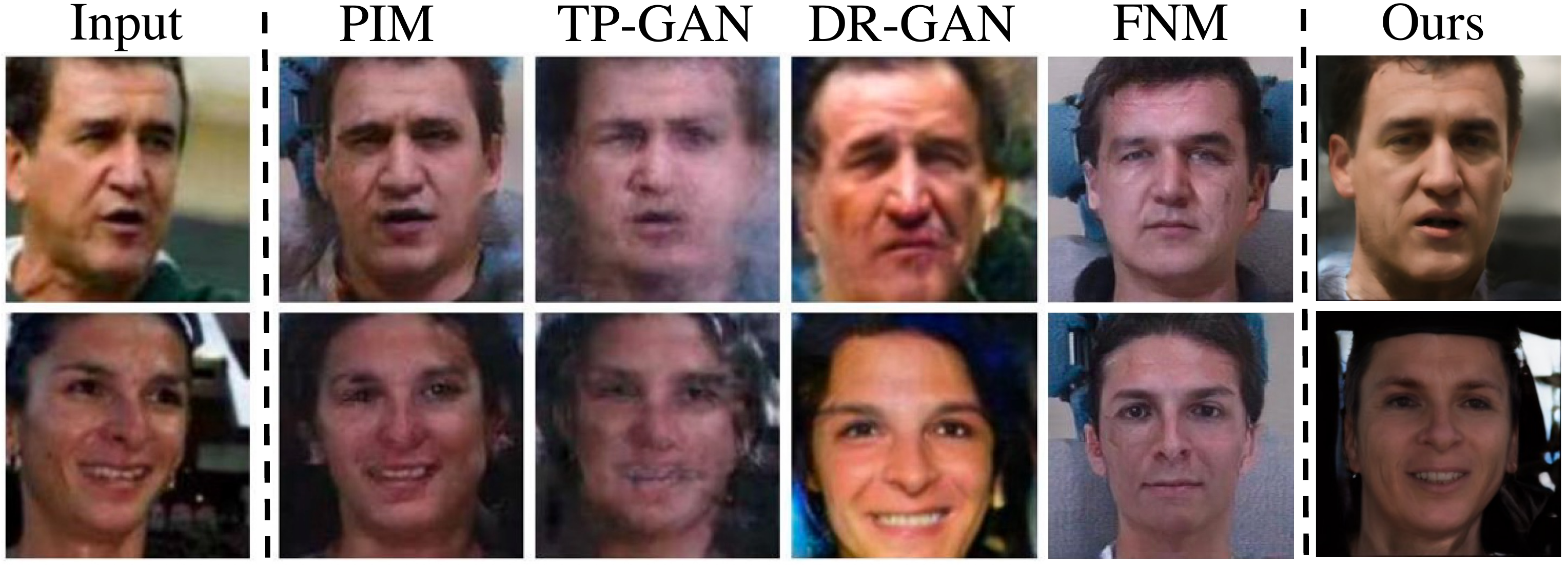}
    \vspace{-0.7cm}
    \caption{Face frontalization on LFW~\cite{LFWTech} images. 
    Our model preserves the best identity with a higher photo realism.
    }
    \label{fig:frontalization}
    \vspace{0.1cm}
    \includegraphics[width=0.5\textwidth]{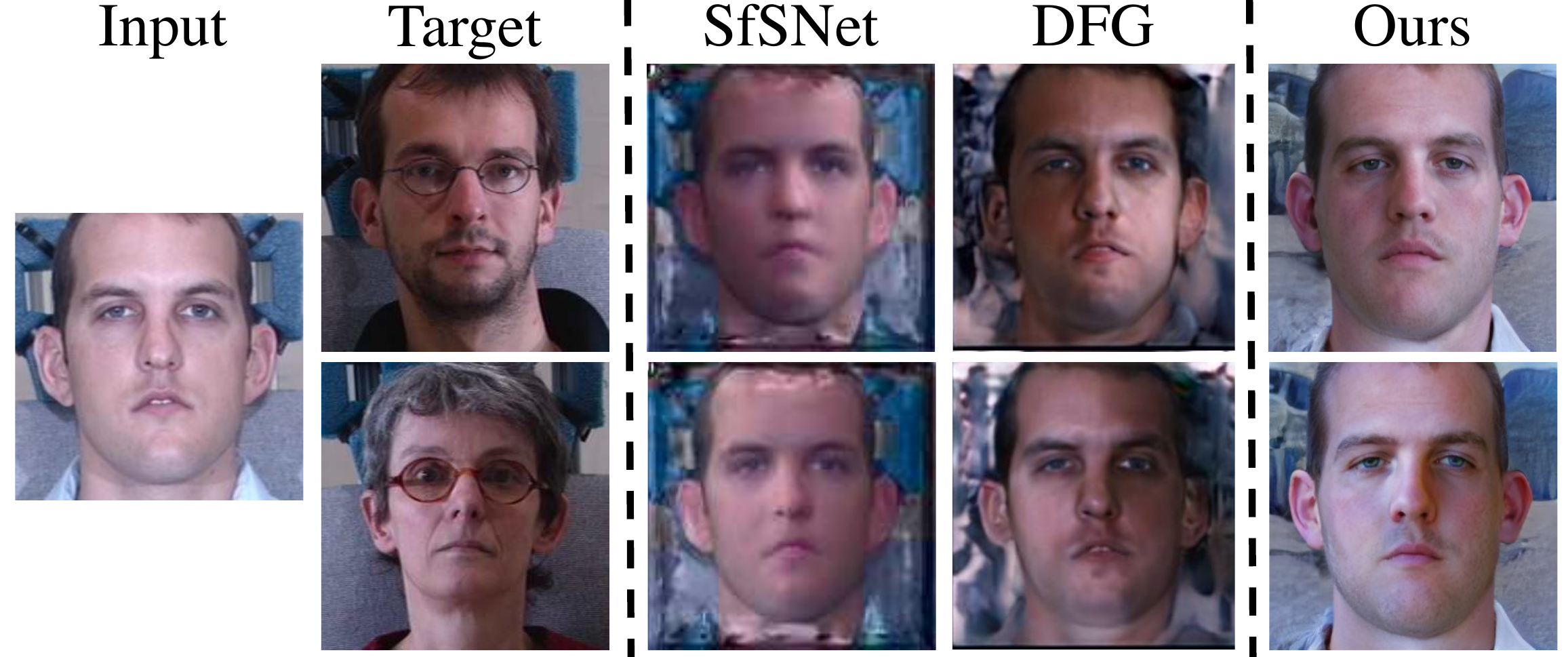}
    \vspace{-0.7cm}
    \caption{Comparing prior arts on portrait relighting with Multi-PIE~\cite{gross2010multi} images.
    Our method provides a higher photo-realism, 
    and merges the indoor light with the person's skin tone more naturally.
    }
    \label{fig:relighting}
    \vspace{-1.3cm}
\end{wrapfigure}

\noindent\textbf{Frontalization.} In Fig.~\ref{fig:frontalization}, we compare our model with prior methods~\cite{zhao2018towards,huang2017beyond,tran2017disentangled,qian2019unsupervised} on the tasks of face frontalization on LFW~\cite{LFWTech},
where
our method best preserves the face identity and produce more photo-realistic images.

% \begin{wrapfigure}{r}{0.6\textwidth}
%     \vspace{-0.7cm}
%     \centering
%     \includegraphics[width=0.6\textwidth]{Figure/pose_edit_comparison_against_SOTA_2.png}
%     \vspace{-0.7cm}
%     \caption{
%     Visual comparisons with CONFIG~\cite{kowalski2020config} on manipulating the same real image and with DiscoFaceGAN's~\cite{deng2020disentangled} generated images.
%     Our model produces better editability at large pose rotations with higher photo-realism and better identity preservation.
%     }
%     \label{fig:pose_generalizability}
%     \vspace{-0.6cm}
% \end{wrapfigure}

\noindent\textbf{Relighting.} 
We show portrait relighting comparison with~\cite{sengupta2018sfsnet,deng2020disentangled} on Multi-PIE~\cite{gross2010multi} in Fig.~\ref{fig:relighting}.
While SfsNet and DFG do not synthesize realistic manipulation with artifacts in background and around the face,
our method shows higher photo-realism and can preserve background pattern like the clothes around the neck.
Moreover, DFG completely changes the skin tone of the person,
whilst our method meshes the extreme indoor light with the skin tone more naturally.

\begin{comment}
We further show visual comparisons with CONFIG and DiscoFaceGAN in Fig.~\ref{fig:pose_generalizability}.
Specifically, we manipulate the same real image with both our method and CONFIG,
and we compare the generated images from DiscoFaceGAN at the same pitch rotations.

Compared with CONFIG, we see that our model preserve better identity at all rotation angles, which is consistent with Fig.~\ref{fig:identity_preservation}.
At large rotation degrees, both prior arts fail to produce photo-realistic images and the identities are also lost.
On the contrary, our model can still produce high-fidelity images with well preserved identities,
which indicates that it enjoys a larger range of editability.
\end{comment}
\section{Ablation Study}

\subsection{Training Strategy}\label{sec:strat_ablation}

In Sec.~\ref{sec:train_strat}, we propose to do alternate training between reconstruction and disentanglement.
To understand its effectiveness, 
we conduct a study and find both of them are essential to learn high-quality identity-preserved editing with results shown in Fig.~\ref{fig:ablation_study} (\textbf{Left}).
Specifically, we perform 140k training iterations with synthetic data on a Render-$\W$ with the following variants: 
(1) reconstruction only training (\textbf{Col.~3});
(2) disentangled only training (\textbf{Col.~4});
(3) alternate training (70k iterations for each) between reconstruction and disetanglement  (\textbf{Col.~5}).
While reconstruction only training enables good identity preservation, it's hard for the model to respond to the editing signals. 
On the other hand, disentangled only training provides good editability, but fails to preserve the identity like face shapes, ages, etc.
Different from them, alternating between these two strategies helps the model achieve a much better performance as it picks up information from both sides.

\subsection{Two-Phase Training}

\begin{figure}[t]
    \centering
    \includegraphics[width = \textwidth]{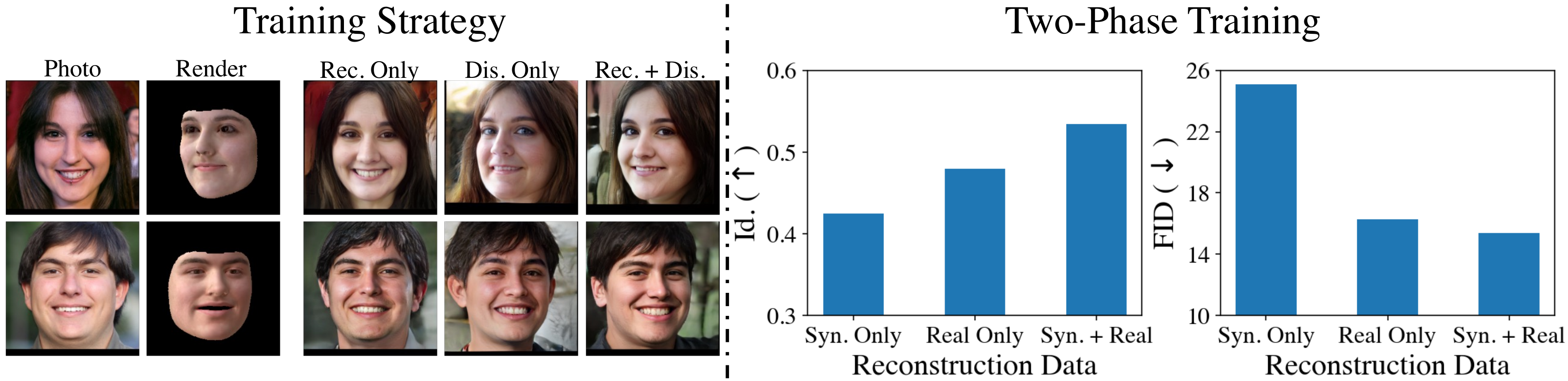}
    \vspace{-0.6cm}
    \caption{Ablation study. \textbf{Left:} 
    Effectiveness of alternate training.
    Compared to reconstruction or disentangled only training, alternate training scheme acquire information for both editing and identity preservation. 
    \textbf{Right:} Effectiveness of two-phase training.
    Using FFHQ for reconstruction significantly improves the photo-realism. 
    The two-phase scheme, fine-tuning with synthetic reconstruction first and then switch to FFHQ,
    further improves identity preservation.
    }
    \label{fig:ablation_study}
    \vspace{-0.7cm}
\end{figure}

% We also propose a two-phase training scheme,
% where the first and the second phase use synthetic and real images for reconstruction training, respectively.
% Note disentangled training is achieved by synthetic data in both phases.
To study our two-phase training scheme, where different data are used for reconstruction training,
we adopt a Render-$\W$ architecture and train for 280k iterations with the following schedules:
(1) synthetic data only reconstruction; 
(2) real data only reconstruction; 
(3) 140K iterations of synthetic reconstruction followed by 140K iterations of real reconstruction.
In Fig.~\ref{fig:ablation_study} (\textbf{Right}), we see that incorporating real data for reconstruction training is crucial for achieving high photo-realism.
Moreover, the two-phase training scheme, (3), yields the best identity preservation.

% \begin{figure}[t]
%     \centering
%     \includegraphics[width=0.47\textwidth]{Figure/Strategy_Ablation_Plot.png}
%     \vspace{-0.3cm}
%     \caption{Effectiveness of alternate training.
%     Compared to reconstruction or disentangled only training, alternate training scheme acquire information for both editing and identity preservation. 
%     }
%     \vspace{-0.4cm}
%     \label{fig:strat_ablation_plot}
% \end{figure}

% \begin{figure}
%     \centering
%     \includegraphics[width = 0.46\textwidth]{Figure/Training_Strategy_Investigation_Bar_3.png}
%     \vspace{-0.3cm}
%     \caption{ 
%     Effectiveness of two-phase training.
%     Using FFHQ for reconstruction significantly improves the photo-realism. 
%     The two-phase scheme, fine-tuning with synthetic reconstruction first and then switch to FFHQ,
%     further improves identity preservation.
%      }
%     \label{fig:training_phases}
%     \vspace{-0.45cm}
% \end{figure}

%\input{limitation}
\section{Conclusion}

In this work, we propose \Ours, a novel framework for high-quality, 3D-controllable, existing face manipulation.
Unlike prior works, 
our model is trained exactly for the task of face manipulation,
and does not require any manual tuning after the learning phase.
We design two training strategies that are both essential for the model to gain abilities of  high-quality, identity-preserved editing.
We also study the information encoding scheme on StyleGAN's latent spaces, 
which leads us to a novel multiplicative co-modulation architecture.
We carry out qualitative and quantitative evaluations on our model,
where it all demonstrates good editability, strong identity preservation and high photo-realism, 
outperforming the state of the arts.
More surprisingly, our model shows a strong generalizability, 
where it can perform controllable editing on out-of-domain artistic faces.

\clearpage
%%%%%%%%% REFERENCES
{\small
\bibliographystyle{Sty_Ref/splncs04}
\bibliography{Sty_Ref/egbib}
}

\newpage
We organize our supplementary as follows.
In Sec.~\ref{sec:implementation}, we include more implementation details.
In Sec.~\ref{sec:extra_ablation}, we show more comparison results among different designs of co-modulation architectures. 
In Sec.~\ref{sec:extra_visual}, we include more results for single factor disentangled editing as well as reference based generation. 
More comparisons with CONFIG~\cite{kowalski2020config} and VariTex~\cite{buehler2021varitex}  are shown in Sec.~\ref{sec:extra_sota_comparison}.
We discuss the limitation and potential societal impacts of our work in Sec.~\ref{sec:limitation}.

\section{Detailed Implementations}\label{sec:implementation}

\subsection{Modules and Training Strategies}

We adopt implementations of the face reconstruction network $\FR$, 
the BFM 3DMM,
and the renderer $\Rd$ all from DiscoFaceGAN~\cite{deng2020disentangled} released repository\footnote{\url{https://github.com/microsoft/DiscoFaceGAN}}.
We use a public implementation\footnote{\url{https://github.com/rosinality/stylegan2-pytorch}} of  
StyleGAN2~\cite{karras2020analyzing} generator and discriminator. 
The ResNet-18~\cite{he2016deep} encoder for $\mathbf{E_T}$ and $\mathbf{E_W}$ is provided by official release\footnote{\url{https://pytorch.org/vision/stable/models.html}} in PyTorch~\cite{paszke2019pytorch}. 
We use the official implementation\footnote{\url{https://github.com/eladrich/pixel2style2pixel}} of PSP encoder~\cite{richardson2021encoding}.

Our StyleGAN generator $\mathbf{G_s}$ and discriminator $\mathbf{D_s}$ are initialized with the pre-trained weights from the unconditional noise-to-image unpaired training regime.
All encoders, $\mathbf{E_T}$, $\mathbf{E_W}$, and $\mathbf{E_{W^+}}$ are initialized randomly.
We adopt two Adam optimizers~\cite{kingma2014adam} to update the parameters in $\G$ ($\mathbf{G_s}$ and $\mathbf{E}$) and $\mathbf{D_s}$ separately.
In phase-1 training, we set our learning rate to be 0.0001 while in phase-2 training, the learning rate is set to be 0.001.

The face recognition network~\cite{deng2019arcface},
the landmark detection model~\cite{bulat2017far},
and the LPIPS module~\cite{richardson2017learning} are from \footnote{\url{https://github.com/ronghuaiyang/arcface-pytorch}}, \footnote{\url{https://github.com/1adrianb/face-alignment}}, and \footnote{\url{https://github.com/richzhang/PerceptualSimilarity}}.

\begin{figure}[t]
    \centering
    \includegraphics[width = \textwidth]{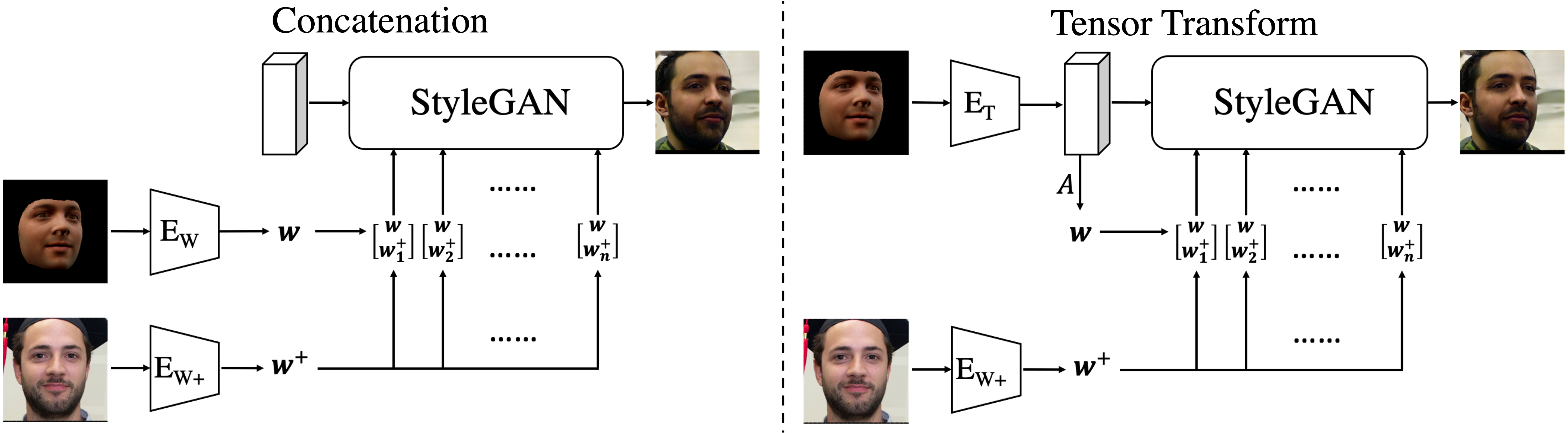}
    \caption{Two types of co-modulation architectures. \textbf{Left:} Concatenation co-modulation architecture.
    \textbf{Left:} Tensor transform co-modulation architecture.
    }
    \vspace{-0.35cm}
    \label{fig:co_mod_architecture}
\end{figure}

\subsection{Co-Modulation Architectures}

In addition to the multiplicative co-modulation architecture,
we also investigate the concatenation and tensor transform co-modulation, 
as shown in Fig.~\ref{fig:co_mod_architecture}. 
For concatenation scheme, 
we encode $R$ by $\mathbf{E_W}$ into the $\W$ space. 
Then, for layer $l$, the modulation signal is provided by concatenating $\W$ and $W^{+}_{l}$ as [$\W, \W^{+}_{l}$]$\in\Real^{1024}$.
The tensor transform scheme originally proposed in~\cite{zhao2021large} is similar to the concatenation scheme in terms of generating the co-modulation signals,
while its $R$ is encoded into $\T$ and an additional linear transformation layer $A$ transforms the flattened $\T$ into $\W$.

\subsection{Evaluation of DiscoFaceGAN}

\textbf{FID.} We follow a similar procedure as Sec.~4.2 of the main paper to generate manipulated images with the generator $\Gd$ from DiscoFaceGAN (DFG)~\cite{deng2020disentangled}. 
The process is similar except that we need to sample an extra noise $n$ for generating each edited image $\hat{P_d}~\text{=}~\Gd(\hat{p}, n)$.
We then measure the FID between $\hat{\P_{d}}$ and $\P$.

\noindent\textbf{Image Manipulation.} 
Since DFG does not provide codes for its image editing, we implement it on our own, strictly following Eqn. 11 of its paper:
(1) Given a photo $P$, obtain its 3DMM parameter by face reconstruction network $p$ = $\FR$($P$), 
and its latent code $w^{+}$ by StyleGAN inversion~\cite{abdal2020image2stylegan++}\footnote{We run 3000 optimization steps to fully retrieve the latent code. 
The implementation is at: \url{https://github.com/Puzer/stylegan-encoder}}.
(2) with the desired manipulation $\hat{p}$, offset $w^{+}$ by $\Delta w(p, \hat{p})$ to generate the manipulated face.

\noindent\textbf{Analysis of $\lambda$ and $\mathcal{W}+$ Space.}
We conduct an analysis of DFG's $\lambda$ and $\mathcal{W}^+$ space in Fig.~13 of the main paper.
Specifically, given a photo $P$, we extract its 3DMM parameter by face reconstruction and $p$ = $\FR$($P$) and do the following for the two spaces: 
(1) $\mathcal{W}+$ space: follow the above image manipulation step with a series of manipulation $\hat{p}$. 
(2) $\lambda$ space: sample a noise $n$ and conduct forward-only inference (no back-propagated optimization) with its generator to synthesize images of the original parameter $\Gd(p, n)$ and from the manipulated parameters $\Gd(\hat{p}, n)$.

\noindent\textbf{Run-Time Efficiency.} 
To manipulate an image, 
DFG would take around 120s to retrieve the latent code on a P100 GPU, 
followed by a 0.5s synthesis process.
On the contrary, our method only takes less than 1s for face reconstruction and around 0.7s for image generation on the same hardware.
Hence, our method enjoys a speedup of 70$\times$ (1.7s vs. 120.5s) for single image editing.

\section{Co-Modulation Comparison}\label{sec:extra_ablation}

\begin{figure}[t]
    \centering
    \includegraphics[width=0.9\textwidth]{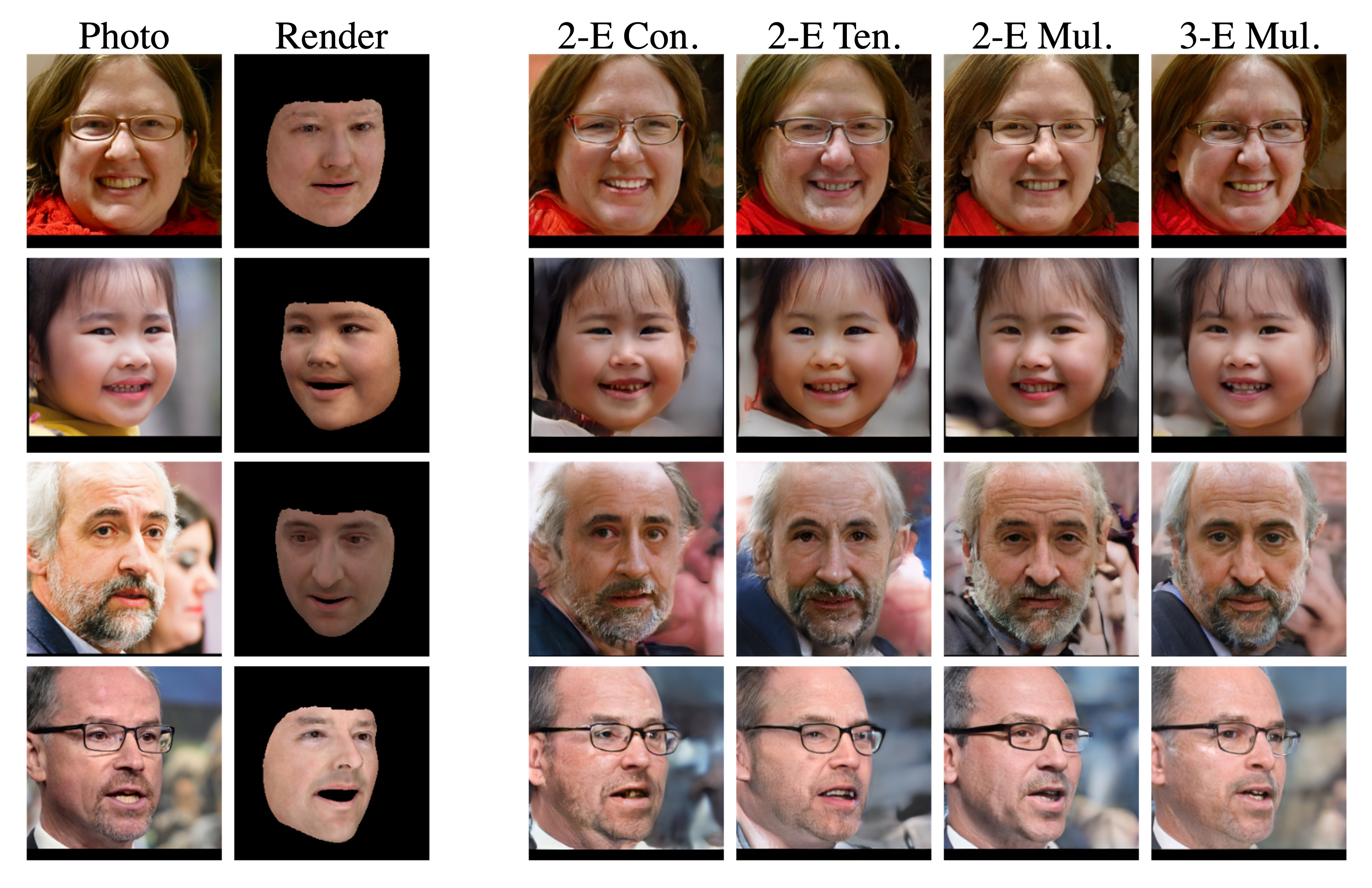}
    \caption{Visual comparison among different co-modulation architectures.
    While concatenation (\textbf{Col.~3}) and tensor transform (\textbf{Col.~4}) schemes have obvious photo-realism issues (\textbf{Row~2~\&~4}),
    the multiplication scheme (\textbf{Col.~5~\&~6}) generally synthesizes images with higher quality, 
    where the 3-encoder architecture (\textbf{Col.~6}) further enhances the editability (\textbf{Row~1~\&~2}).
    }
    \label{fig:co_mod_ablation}
    \vspace{-0.35cm}
\end{figure}

In addition to the quantitative results in Tab.~1 of the main paper, 
we further show visual comparisons of the 4 proposed co-modulation schemes in Fig.~\ref{fig:co_mod_ablation},
where we compare 2-encoder concatenation (\textbf{Col.~3}),
2-encoder tensor transform (\textbf{Col.~4}),
2-encoder multiplication (\textbf{Col.~5}), and 3-encoder multiplication (\textbf{Col.~6}) co-modulation architectures.
As shown in the figure, 
the concatenation and tensor transform schemes would have photo-realism issues with significant amount of artifacts (\textbf{Row~4}) and unrealistic poses (\textbf{Row~2}).
On the contrary, the multiplicative scheme performs much better,
and the 3-encoder multiplicative co-modulation further demonstrates better editability in lighting (\textbf{Row~1}) and pose (\textbf{Row~2}).

\section{Additional Image Manipulations}\label{sec:extra_visual}

We show more results of disentangled editing in Fig.~\ref{fig:sig_fac_dis_editing}.
Our model again provides highly disentangled manipulation with high photo-realism and strong identity preservation.
We also show additional reference based face generation results with more identities in Fig.~\ref{fig:ref_face_gen}.
Noticeably, even the identity images with extreme poses (\textbf{2nd and 8th identites}) can be well re-posed with the expression and illumination transferred.
Our model also well preserves the eyeglasses in manipulating the 7th identity image.

\begin{figure}[t]
\centering
\includegraphics[width=0.92\textwidth]{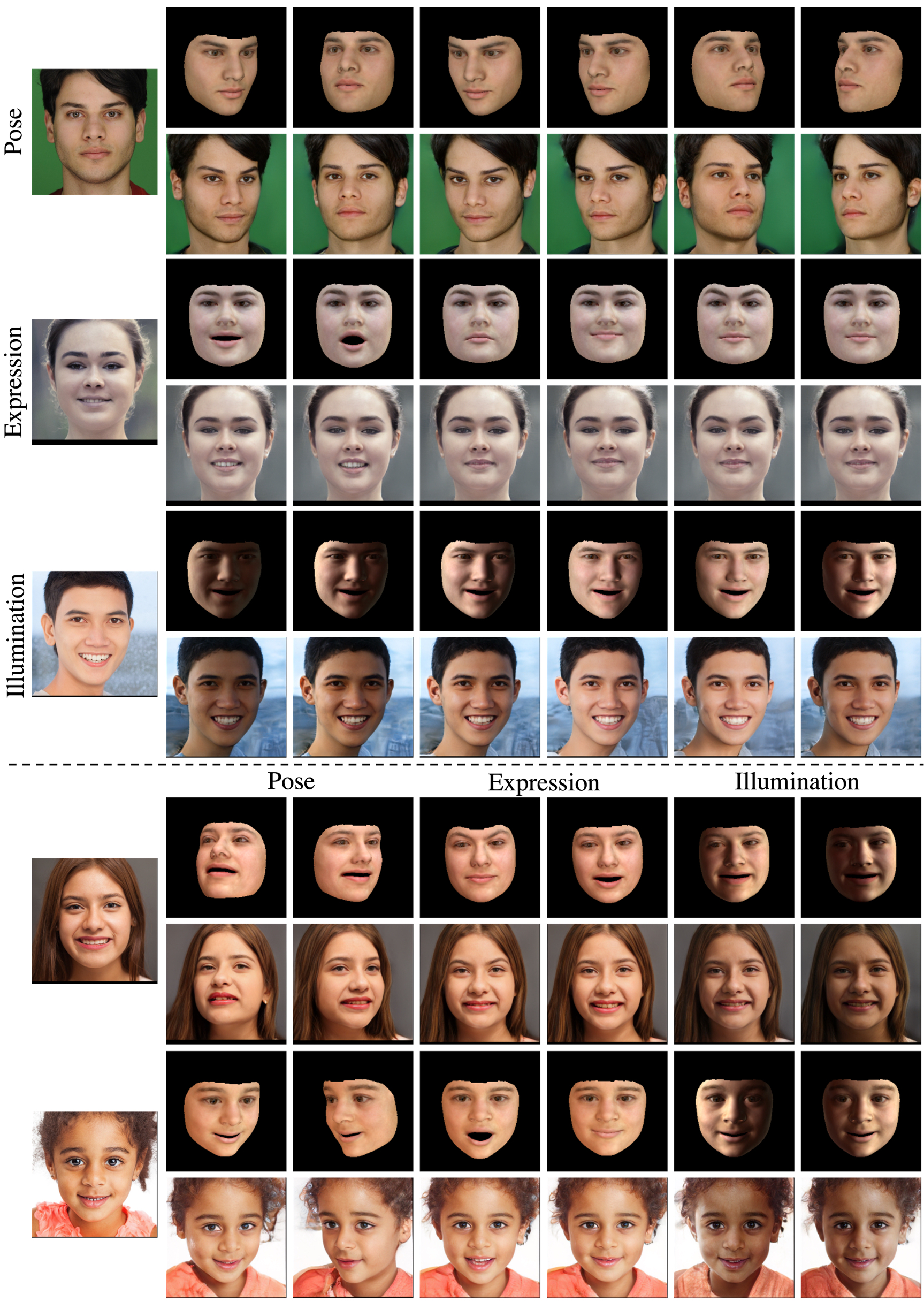}
    \caption{More results for disentangled editing.
    Our model again achieves good disentangled editability, high photo-realism, and strong identity preservation.}
    \label{fig:sig_fac_dis_editing}
\end{figure}

\begin{figure}[t]
    \centering
    \includegraphics[width=1.0\textwidth]{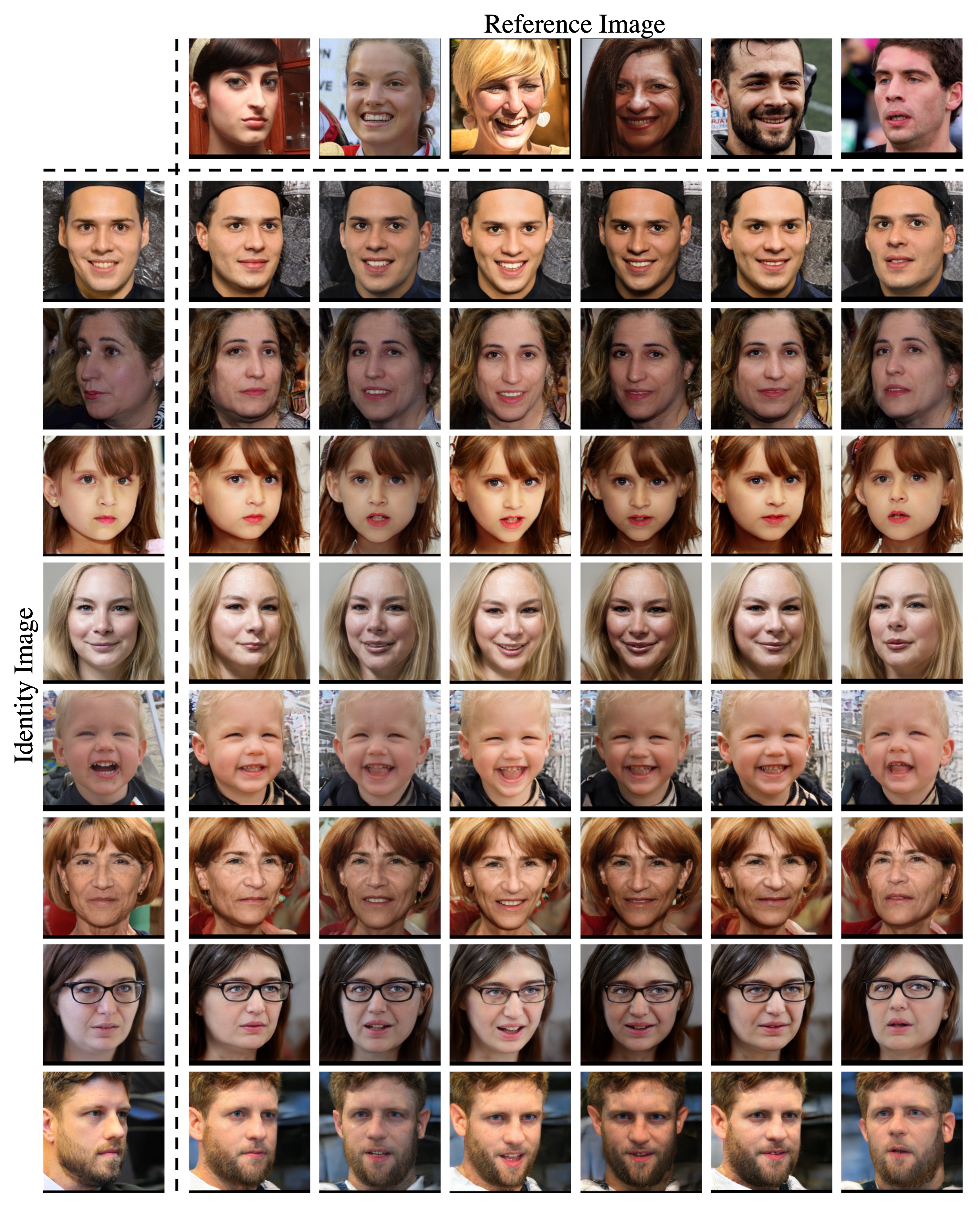}
    \caption{Additional examples for reference based face generation.
    Our model demonstrates a good editability with identities with extreme poses (\textbf{2nd and 8th}) and with eyeglasses (\textbf{7th}).
    }
    \label{fig:ref_face_gen}
\end{figure}

\section{More Comparisons to SOTA Methods}\label{sec:extra_sota_comparison}

\begin{figure}[t]
    \centering
    \includegraphics[width=1\textwidth]{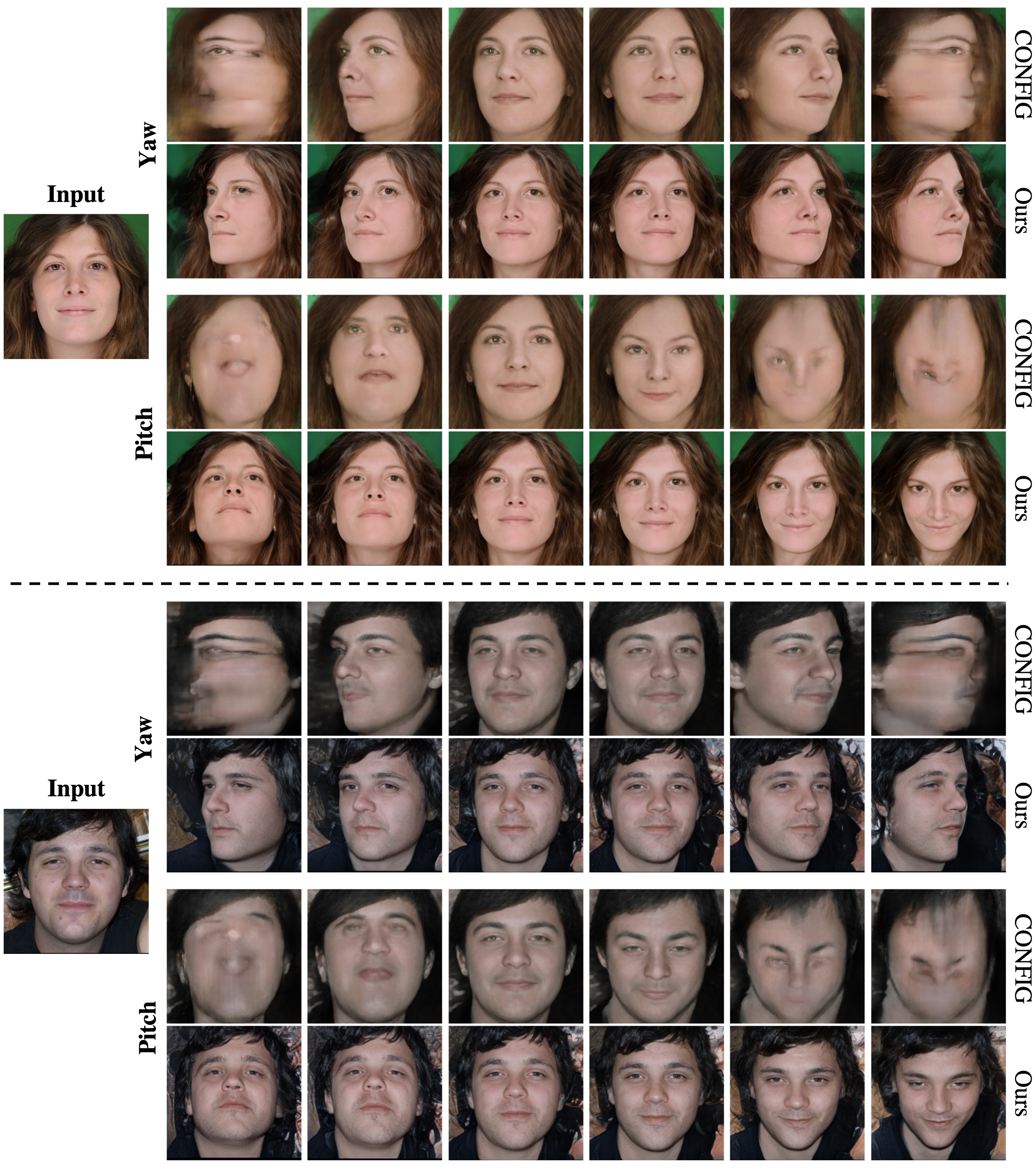}
    \vspace{-0.35cm}
    \caption{More comparison with CONFIG~\cite{kowalski2020config} on real image editing.
    Our method again outperforms CONFIG with larger range of editability,
stronger identity preservation, and higher photo-realism.
    }
    \label{fig:config_comparison}
\end{figure}

We show additional comparisons with CONFIG~\cite{kowalski2020config} in Fig.~\ref{fig:config_comparison} on both yaw and pitch rotations for real image.
Clearly seen from the plot again, our method enjoys larger range of editability,
stronger identity preservation, and higher photo-realism.

\begin{figure}[t]
    \centering
    \includegraphics[width=1\textwidth]{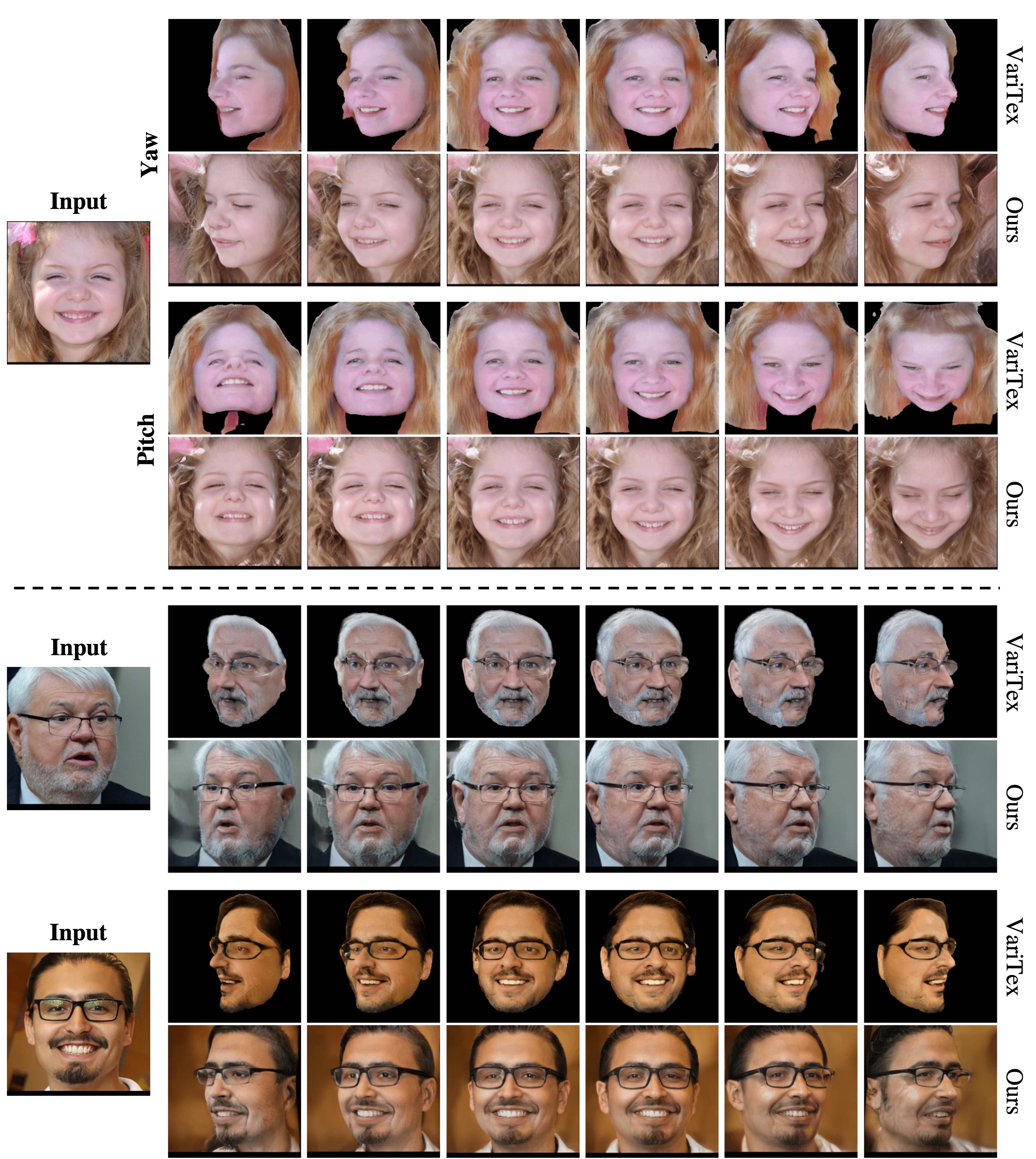}
    \vspace{-0.35cm}
      \caption{Manipulating the same real images with our model and VariTex~\cite{buehler2021varitex},
      where our model shows better identity preservation in all examples.
    \textbf{Top:} While VariTex could not synthesize realistic hair and background due to the absence of 3DMM modelling, 
    our model provides a much better synthesis result on these regions, demonstrating a higher photo-realism and better editability. 
    \textbf{Bottom:}
   We manipulate faces with additional rigid bodies like glasses.
    Our method again generates images with much higher photo-realism, even at extreme poses. 
    }
    \label{fig:varitex_comparison}
\end{figure}

We further include comparisons with VariTex~\cite{buehler2021varitex} in Fig.~\ref{fig:varitex_comparison} on real image manipulation.
Again, we find a clear advantage in identity preservation for our model over VariTex for all inputs.
As shown in the \textbf{Top} example, while VariTex could not synthesize background and the generated hair has a unrealistic texture,
our method demonstrates a much higher photo-realism with better background and hair.
Moreover, we compare the editability between our model and VariTex by manipulating images with extra rigid bodies, the eyeglasses, which represents a harder task in the \textbf{Bottom} example.
While VariTex could not properly synthesize the faces with glasses, 
our model provide a decent control to generate high-quality images.

\section{Limitations \& Societal Impacts}\label{sec:limitation}

Although \Ours\ shows a strong ability for 3D-controllable, identity-preserved face editing, 
there remains certain limitations and potential negative impacts. 

\noindent\textbf{Limitations.} Our 3DMMs can not model fine details like wrinkles and hair styles,
and thus our model can not explicitly control those attributes.
We also see a gap between the reconstructed images and the inputs in face shapes, which might be caused by the imprecise 3D estimation in face reconstruction.
Moreover, our model would inherit bias from the training data, and due to lack of public availability,
we can only use synthetic data for disentangled learning.

\noindent\textbf{Potential Negative Impacts.}
Face manipulation techniques has in the past helped creating deep-fakes and spread disinformation. Our work is intended for intelligent content creation for portrait photography and we believe it does not improve the accessibility of deep-fakes and disinformation. Moreover, our discoveries of identity-editability trade-off might also offer new viewpoints on future development for deep-fake detection techniques.

\end{document}